\documentclass[10pt,twocolumn,letterpaper]{article}

\usepackage{cvpr}              
\usepackage{adjustbox}
\usepackage{multirow, makecell}
\usepackage{pifont}

\definecolor{cvprblue}{rgb}{0.21,0.49,0.74}
\usepackage[pagebackref,breaklinks,colorlinks,allcolors=cvprblue]{hyperref}

\def\paperID{9899} 
\def\confName{CVPR}
\def\confYear{2026}

\title{Attribution as Retrieval: Model-Agnostic AI-Generated Image Attribution}

\author{
	Hongsong Wang\textsuperscript{1,2}, Renxi Cheng\textsuperscript{3}, Chaolei Han$^{3}$, Jie Gui\textsuperscript{3,4,5}\thanks{Corresponding author}\\
	$^{1}$School of Computer Science and Engineering, Southeast University, Nanjing 210096, China \\
	$^2$Key Laboratory of New Generation Artificial Intelligence Technology and Its Interdisciplinary \\
	Applications (Southeast University), Ministry of Education, China \\ 
	$^{3}$School of Cyber Science and Engineering, Southeast University, Nanjing 210096, China\\
	$^{4}$Purple Mountain Laboratories, Nanjing 210000, China \\
	$^{5}$Engineering Research Center of Blockchain Application, Supervision And Management\\ (Southeast University), Ministry of Education, China \\
	\tt\small\{hongsongwang, renxi, chaoleihan, guijie\}@seu.edu.cn \\
}

\begin{document}
\maketitle

\begin{abstract}
With the rapid advancement of AIGC technologies, image forensics will encounter unprecedented challenges. Traditional methods are incapable of dealing with increasingly realistic images generated by rapidly evolving image generation techniques. To facilitate the identification of AI-generated images and the attribution of their source models, generative image watermarking and AI-generated image attribution have emerged as key research focuses in recent years. However, existing methods are model-dependent, requiring access to the generative models and lacking generality and scalability to new and unseen generators. To address these limitations, this work presents a new paradigm for AI-generated image attribution by formulating it as an instance retrieval problem instead of a conventional image classification problem. We propose an efficient model-agnostic framework, called Low-bIt-plane-based Deepfake Attribution (LIDA). The input to LIDA is produced by Low-Bit Fingerprint Generation module, while the training involves Unsupervised Pre-Training followed by subsequent Few-Shot Attribution Adaptation.
Comprehensive experiments demonstrate that LIDA achieves state-of-the-art performance for both Deepfake detection and image attribution under zero- and few-shot settings. The code is at \url{https://github.com/hongsong-wang/LIDA}
\end{abstract}

\section{Introduction} \label{sec:intro}
With the rapid advancement of AI-Generated Content (AIGC) technologies, such as image generation~\cite{bie2024renaissance}, motion generation~\cite{wang2026temporal,weng2025realign,tan2026easytune,tansopo} and video generation~\cite{11106267}, synthetic media has become increasingly realistic and widely accessible. While AIGC brings significant benefits to entertainment and productivity, it also raises critical concerns regarding authenticity and potential misuse~\cite{zhang2025security}. As a result, AIGC forensics~\cite{xu2025advancements} has emerged as an essential research area, aiming to detect, attribute, and trace AI-generated or AI-manipulated content. Reliable AIGC forensics techniques are crucial for safeguarding digital media integrity and preventing malicious abuse in the era of generative AI.

\begin{figure}[t]
    \centering
    \includegraphics[width=\linewidth]{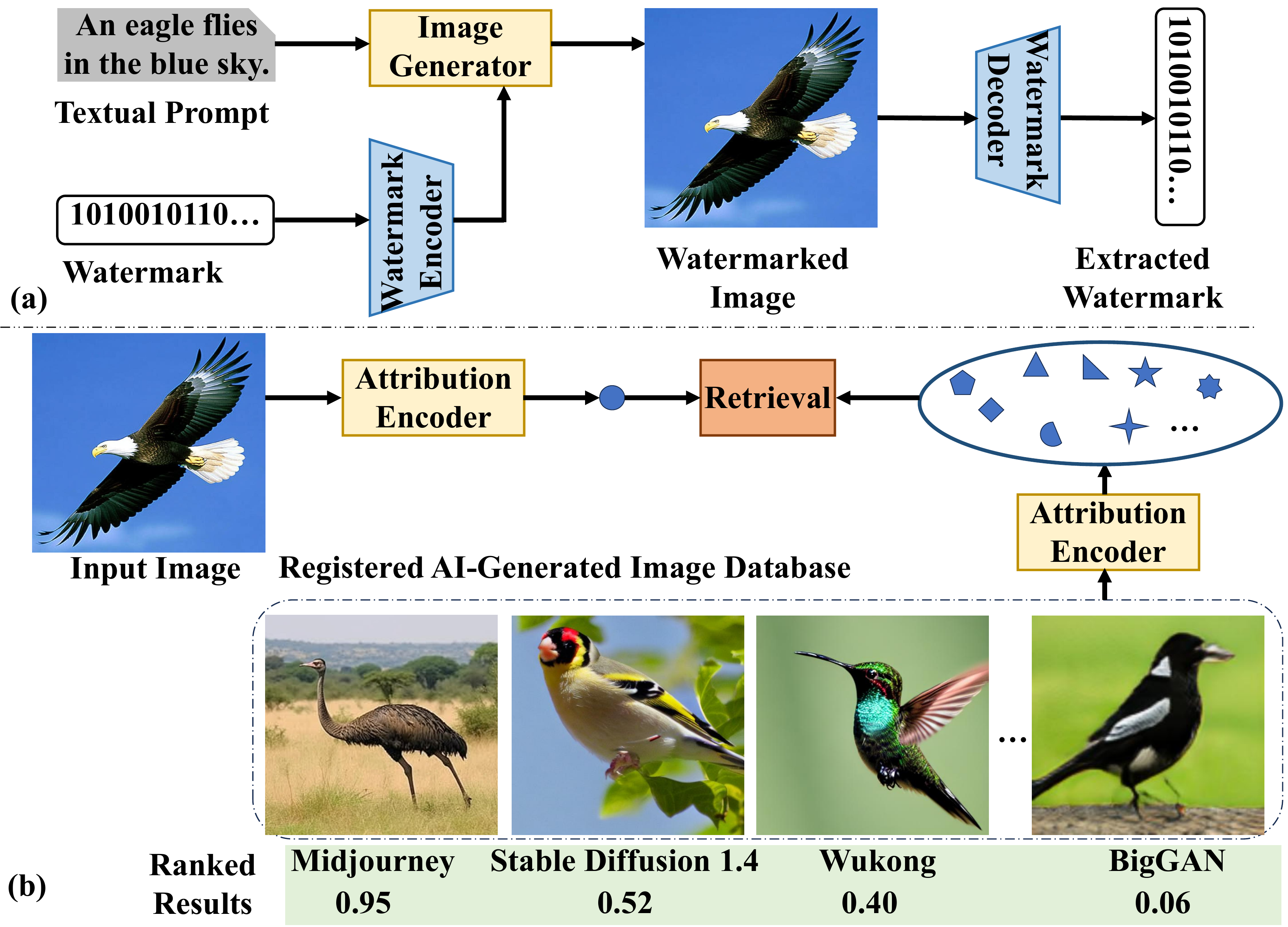}
    \caption{\textbf{Comparison between generative image watermarking and our retrieval-based AI-generated image attribution.} Our framework is concise, versatile, and easily adapted to new image generators.} 
    \label{fig:intro}
\end{figure}

Traditional media forensics methods~\cite{rocha2011vision} struggle to adapt to the challenges posed by AIGC, as AI-generated image or video does not contain camera-based physical traces and exhibits far fewer inconsistency artifacts that conventional forensics rely on. However, although modern generative models can produce highly realistic content, they still leave distinctive generative fingerprints or traces that differ from those found in natural images~\cite{yu2021artificial,song2024manifpt}.

Recently, the detection of AI-generated images has become an increasingly important problem. The research focus has shifted from early approaches that relied on spotting visible artifacts to more robust strategies that emphasize generalization across diverse and unseen generation models~\cite{yan2024transcending,guillaro2025bias}. A number of generator-agnostic methods~\cite{guillaro2025bias,wang2025lota,jia2025secret} have been proposed, which are capable of accurately and efficiently distinguishing real images from fake ones. However, Deepfake or AI-generated image detection only determines whether an image is real or fake, without providing any additional forensic information.

To enable the attribution of AI-generated images, two main research directions have emerged: generative image watermarking~\cite{zhang2024watermarks,gunnundetectable} and AI-generated image attribution~\cite{yu2019attributing,girish2021towards}. The former embeds invisible watermarks into the image during the generation process, while the latter is independent of the image generation step. Although approaches of generative image watermarking achieve high accuracy in attributing AI-generated images, they require full access to the image generation model and often do not generalize across different generators (see Figure~\ref{fig:intro}(a)).

Most studies on AI-generated image attribution target the close-set scenario~\cite{yu2019attributing,bui2022repmix,yang2022deepfake}, which is less applicable to modern generative models that are rapidly evolving. Few works~\cite{girish2021towards,yang2023progressive,sun2023contrastive} address the open-set scenario. However, all existing works treat AI-generated image attribution as a classification problem and require labeled or unlabeled AI-generated images from different generators during training. Therefore, these approaches are not flexible enough to adapt to numerous new or unseen image generators.

To address the above limitations, we study AI-generated image attribution from the novel perspective of instance retrieval, and introduce a model-agnostic framework. As illustrated in Figure~\ref{fig:intro}(b), this framework requires only the training of an attribution encoder and is readily scalable to unseen image generators. To guarantee retrieval-based attribution, a registered AI-generated image database is maintained, containing only a few images for each generator.

More specifically, we introduce a method called LIDA, which consists low-bit fingerprint generation, unsupervised pre-training and few-shot attribution adaptation. Low bit-planes of each RGB channel are used to compose the fingerprint image. During unsupervised pre-training, a pretext task with a corresponding side loss is employed to train a lightweight network on large-scale real images to enhance generalization. Few-shot attribution adaptation uses only a limited number of AI-generated images from the registered dataset, along with an equal number of real images. The adaptation is supervised by the image attribution loss and the Deepfake detection loss. 
Comprehensive evaluations on the GenImage~\cite{zhu2023genimage} dataset and WildFake~\cite{hong2025wildfake} dataset, covering zero-shot and few-shot Deepfake detection as well as cross-architecture and cross-generator image attribution, are conducted to validate the effectiveness and robustness of our bit-plane-based forensic technique.

Our contributions are summarized as follows: 
\begin{itemize}[itemsep=1pt]
\item \textbf{Novel solution for AI-generated image attribution: }We formulate AI-generated image attribution as an instance retrieval problem and address it using bit-planes.
\item \textbf{Versatile and efficient pipeline design: }We propose a simple yet effective pipeline consisting of three modules: low-bit fingerprint generation, unsupervised pre-training and few-shot attribution adaptation.
\item \textbf{Superior zero- and few-shot image forensics results: }Our method achieves state-of-the-art performance on two popular AI-generated image datasets for zero- and few-shot detection and attribution.
\end{itemize}

\section{Related Work}
\label{sec:related_work}

\noindent\textbf{Generative Image Watermarking:}
Generative image watermarking~\cite{zhang2024watermarks,gunnundetectable} is a technique that embeds identifiable signatures into images produced by generative models, enabling the attribution of AI-generated images. Traditional image watermarking protects the copyright of individual images, whereas generative image watermarking safeguards the copyright of the generative model itself. To embed watermarks, existing methods either fine-tune the image decoder of generative models~\cite{fernandez2023stable} or modify their latent representations~\cite{rezaei2024lawa}. For example, Tree-Ring~\cite{wen2023tree} embeds an invisible watermark into the initial noise vector of a diffusion model in Fourier space. Gaussian Shading~\cite{yang2024gaussian} enables plug-and-play watermarking by embedding watermarks into the diffusion model’s latent representations following a Gaussian distribution. Although watermarking techniques achieve high accuracy in attributing AI-generated images, they require access to the generative models and involve modifying them, which limits the flexibility of these methods.

\noindent\textbf{Closed-Set AI-Generated Image Attribution:}
Closed-set AI-generated image attribution is a supervised image classification task that assumes all image generators are known during training. Yu et al.~\cite{yu2019attributing} present a systematic study showing that images generated by Generative Adversarial Networks (GANs) carry distinct model fingerprints that enable attribution. 
Frank et al.~\cite{frank2020leveraging} demonstrate that GAN‑generated images produce consistent artifacts in the frequency domain, which can also be used for source identification. RepMix~\cite{bui2022repmix} attributes generated images to their GAN architecture regardless of semantic content and under benign transformations. Yang et al.~\cite{yang2022deepfake} show that even when GANs are fine‑tuned or retrained, the underlying architecture leaves globally consistent fingerprints that enable attribution. These closed-set approaches, which focus on GAN-based image generators, are less practical and less applicable to modern generative models.

\begin{figure*}[t]
    \centering
    \includegraphics[width=\linewidth]{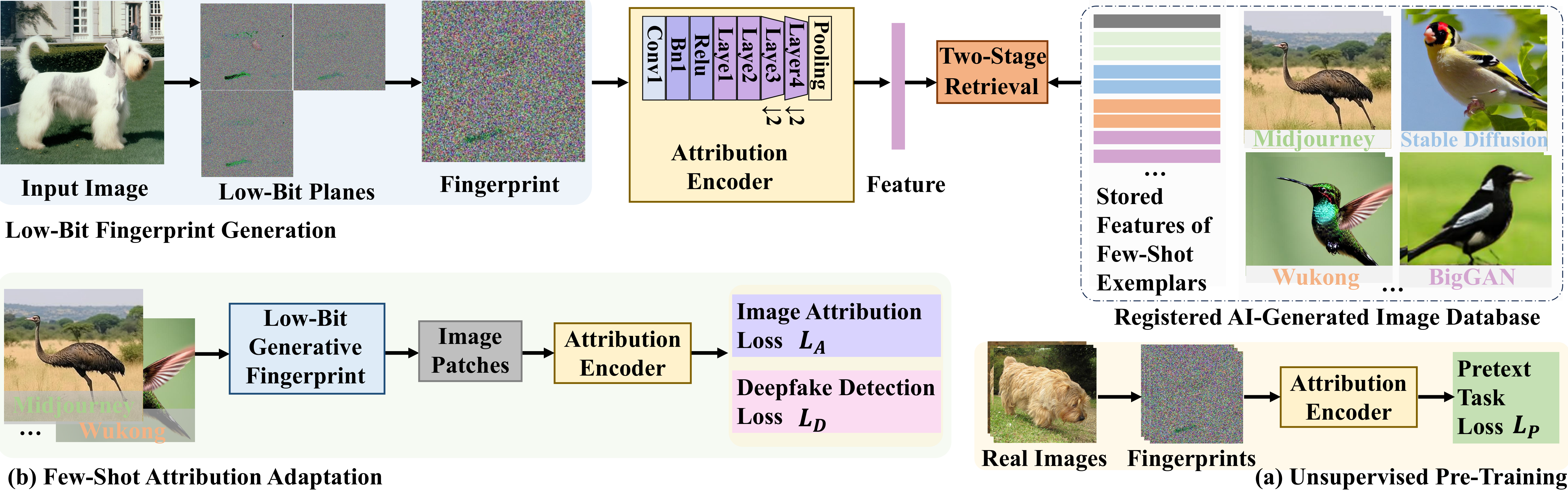}
    \caption{\textbf{Pipeline of the proposed model-agnostic framework LIDA for AI-generated image attribution.} LIDA treats image attribution as an instance retrieval problem, and uses low-bit-plane-based generative fingerprint as the input. The training stage consists of two consecutive steps: (a) unsupervised pre-training and (b) few-shot attribution adaptation.}
    \label{fig:method}
\end{figure*}

\begin{figure}[t]
    \centering
    \includegraphics[width=\linewidth]{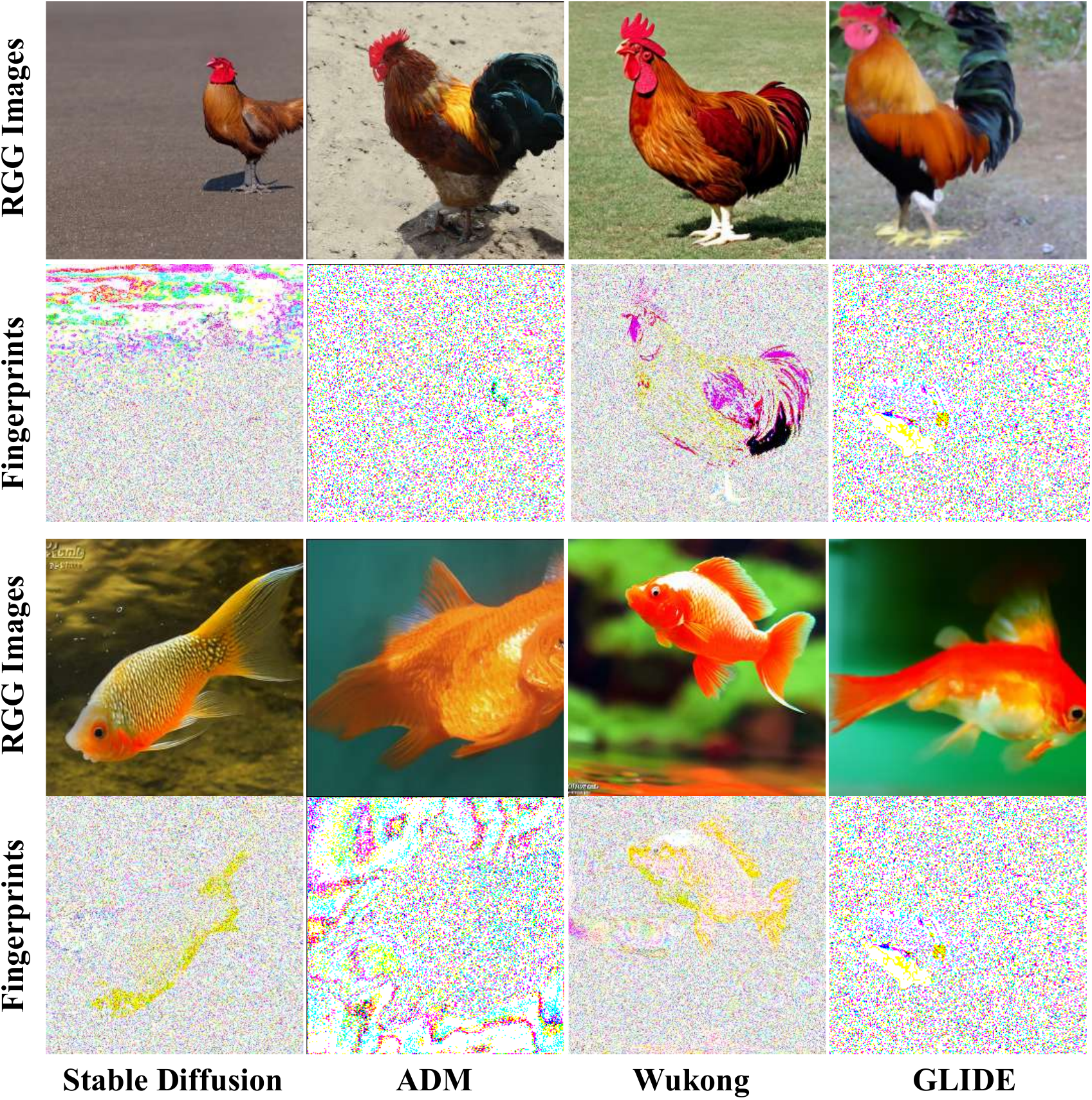}
    \caption{\textbf{Comparison of AI-generated images and low-bit generative fingerprints from different image generators.} Generators include Stable Diffusion~\cite{rombach2022high}, ADM~\cite{dhariwal2021diffusion}, and Wukong~\cite{wukong2022}.}
    \label{fig:fingerpint}
\end{figure}

\noindent\textbf{Open-Set AI-Generated Image Attribution:}
Open-set approaches aim to address the attribution of images generated by new and unseen models. Girish et al.~\cite{girish2021towards} present an algorithm that discovers images generated by unseen GANs and simultaneously attributes them to source models. Yang et al.~\cite{yang2023progressive} simulate open-set samples via lightweight models to enable attribution of both known and unknown generative models. Sun et al.~\cite{sun2023contrastive} combine a voting module and confidence-based pseudo‐labels to attribute forged faces in an open-world scenario. De-fake~\cite{sha2023fake} attributes images from various text-to-image generation models. Li et al.~\cite{li2024handcrafted} show that high-pass handcrafted filters improve attribution performance for both closed‐set and open‐set scenarios. These approaches also treat attribution as a classification task and leverage both labeled and unlabeled images during training, where unseen classes exist in the unlabeled set. However, this setting is still not flexible enough in real-world scenarios, as it requires a large number of unlabeled AI-generated images from new generators for training. Accordingly, this work aims to devise a more flexible and practical paradigm for open-set AI-generated image attribution.

\section{Retrieval Perspective for Image Attribution}
Rather than treating AI-generated image attribution as a classification task, we formulate it as a retrieval task, which naturally supports open-set scenarios by requiring only a well-trained image-based feature extractor for attribution. 

A registered database of AI-generated images, $\mathcal{D} = \{x^1_1, \ldots, x^j_i, \ldots, x^J_N\}$, needs to be maintained, where $j$ denotes the index of the image generator $G_j$ and $i$ denotes the index of the image. To accommodate new image generators, only one or a few example images need to be added to the registered database.

A feature encoder $f(\cdot)$ maps both query and database images into a unified feature space. Given a query image $q$ generated by an arbitrary image generator, the similarity between $q$ and $x^j_i$ is measured using a similarity function based on their extracted features. The most similar images in the registered database are then retrieved.

The retrieval model ranks all database images according to similarity scores, and the top-K retrieved neighbors are defined as:
\begin{equation}
    \text{Top-K}(q) = \operatorname*{arg\,top-K}_{x_i \in \mathcal{D}} \,\mathrm{sim}(q, x_i).
\end{equation}

The attribution decision is based on the generator labels of the retrieved neighbors, e.g., by assigning the label of the top-ranked retrieved image to the query. The main focus is to train or fine-tune the feature encoder $f(\cdot)$, specifically designed for image attribution.

This retrieval-based attribution paradigm eliminates the need for retraining when encountering new generators, making it inherently open-set friendly, as the model can directly incorporate samples from unseen generators into the registered database. This paradigm also provides evidence-based attribution, as the retrieved images justify the predicted source. 

\section{Low-Bit-Plane-Based Deepfake Attribution}
We study AI-generated image attribution (i.e., Deepfake attribution) from the bit-plane perspective, and present a model-agnostic approach called Low-bIt-plane-based Deepfake Attribution (LIDA). LIDA does not need to access models of any image generators. Given only a few AI-generated images as exemplars, it can quickly gain the ability to predict the corresponding image generator of an arbitrary image. The pipeline of LIDA is shown in Figure~\ref{fig:method}. 
Details are described as follows.

\subsection{Low-Bit Fingerprint Generation}
Similar to works on camera fingerprints~\cite{lukas2006digital,chen2008determining}, recent studies demonstrate that AI-generated images also contain distinctive generative fingerprints that enable model attribution and source tracing~\cite{yu2019attributing,yu2021artificial}. Generative fingerprints refer to inherent and consistent artifacts unintentionally embedded by a generative model during the image synthesis process. These artifacts are model-specific and remain stable regardless of the image content.

The low-bit-plane-based AI-generated image detection method~\cite{wang2025lota,wang2026raid} demonstrates that low-bit planes inherently contain intrinsic artifacts that can be exploited to distinguish real from AI-generated images. Motivated by this observation, we hypothesize that such low-bit-plane noise images can also be leveraged for AI-generated image attribution. Thus, we term such noise image low-bit generative fingerprint, which can be quickly obtained via the following simple operations.

For an RGB image $\mathbf{x} \in \mathbb{R}^{H \times W \times 3}$, let $\mathbf{x}_c(i,j)$ denotes the pixel value at position ($i$,$j$) in channel $c \in \{R,G,B\}$. The bit-plane decomposition for each channel is:
\begin{equation}
    \mathbf{x}_c = \sum_{k=0}^{7} 2^k \cdot \mathbf{b}_c^k
\end{equation}
where $\mathbf{b}_c^k$ represents the $k$-th bit plane of the $c$-th channel.

We combine the three least significant bit-planes of each channel and employ the thresholding strategy~\cite{wang2025lota} to construct the generative fingerprint $\mathbf{\tilde{x}}_c$:
\begin{equation}
    \mathbf{\tilde{x}}_c = 255 \cdot \mathrm{sgn}(\sum_{k=0}^{2} 2^k \cdot \mathbf{b}_c^k)
\label{eq:finger_2}
\end{equation}
where $\text{sgn}(\cdot)$ is the sign function which maps elements greater than zero to one, and elements equal to zero to zero.

\begin{figure}[t]
    \centering
    \includegraphics[width=\linewidth]{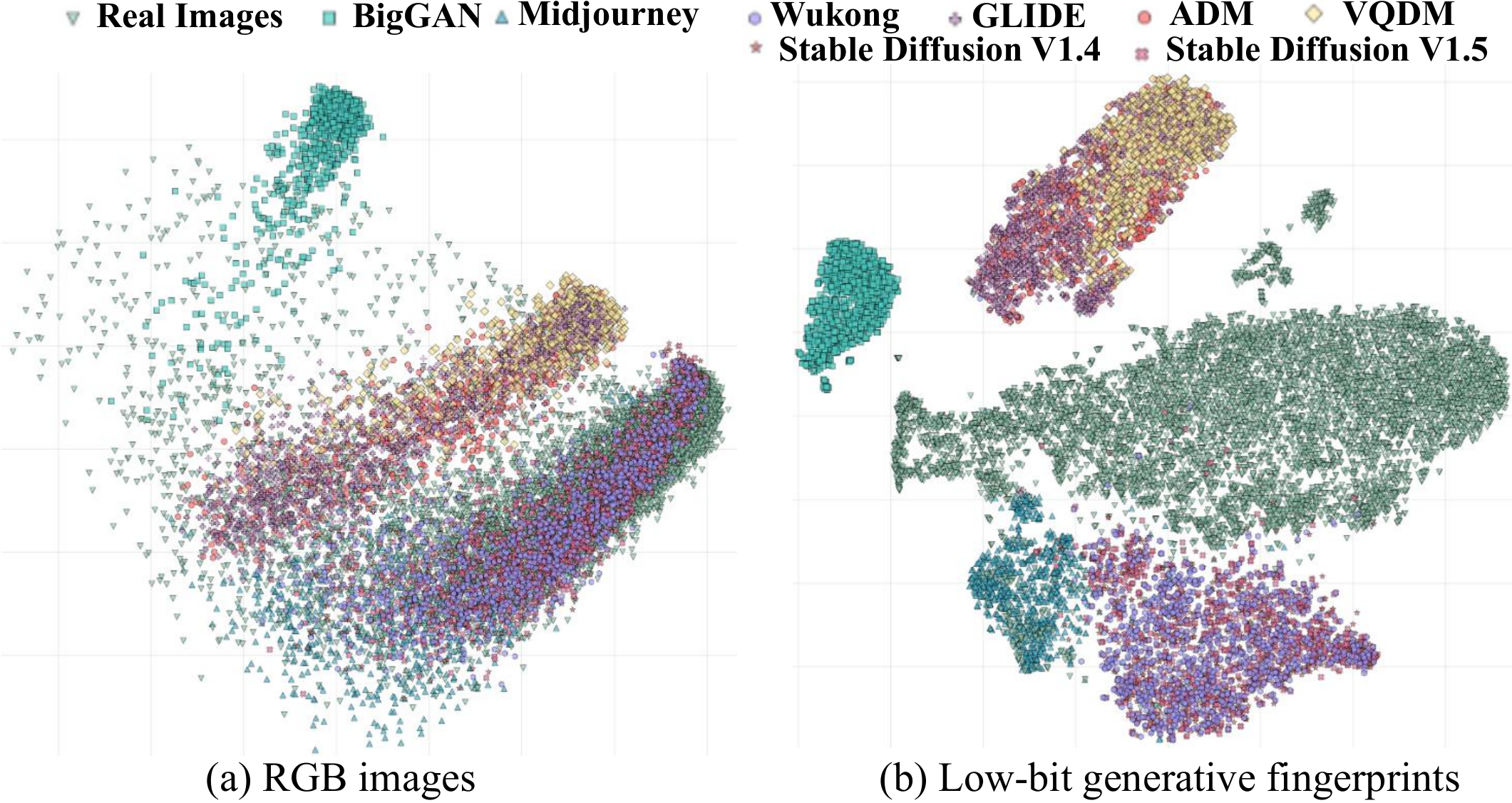}
    \caption{\textbf{Feature distribution of images from different sources} for (a) RGB images and (b) low-bit generative fingerprints.}
    \label{fig:feat_pca}
\end{figure}

We visualize in Figure~\ref{fig:fingerpint} the low-bit generative fingerprints of AI-generated images with the same content, produced by different image generators, including Stable Diffusion~\cite{rombach2022high}, ADM~\cite{dhariwal2021diffusion}, Wukong~\cite{wukong2022}, and GLIDE~\cite{nichol2021glide}. Compared with the original RGB images, the low-bit generative fingerprints, which discard most of the image content, better reveal distinctive traits for model attribution.

To further analyze the distinctive capabilities of the low-bit generative fingerprint, we directly extract features using a pretrained ResNet-50~\cite{he2016deep} on ImageNet~\cite{deng2009imagenet}, and visualize the PCA-reduced features of thousands of samples from different image generators in Figure~\ref{fig:feat_pca}. For RGB images, real images are mixed with AI-generated images, and the distribution differences among images from different generators are almost indistinguishable. In contrast, for low-bit generative fingerprints, real and AI-generated images are clearly separated, and images from the same generator are relatively clustered. Therefore, we use low-bit generative fingerprints as input to train the network to learn features for AI-generated image attribution.

\subsection{Unsupervised Pre-Training}
To enhance generalization, we adopt unsupervised pre-training on a large-scale real image dataset using the fingerprint input computed in Eq.~(\ref{eq:finger_2}). After pre-training, the network learns to capture intrinsic noise structures that are transferable to downstream tasks of generative image forensics. Moreover, the unsupervised pre-training provides a robust weight initialization, leading to faster convergence and improved performance during fine-tuning.

For both simplicity and effectiveness, we adopt ResNet-50~\cite{he2016deep} as the attribution encoder. To better preserve spatial information, we modify the network by removing the downsampling operations in the lower layers. This design maintains high-resolution feature maps, which are essential for capturing subtle structural details in forensic analysis.

We use the pretext task training strategy, and train the attribution encoder on the ImageNet~\cite{deng2009imagenet}. As an example, the image classification is used as the side task to train the network. The pretext task training loss $\mathcal{L}_P$ is formulated as:
\begin{equation}
    \mathcal{L}_P = - \sum_{b=1}^{B} \sum_{c=1}^{C} s_{b}^c \log q_{b}^c,
\end{equation}
where $b$ is the index of fingerprints of real images, $q_{b}$ is the corresponding ImageNet category label, $s_{b}$ is the predicted class probabilities, $c$ is the category index, $B$ and $C$ represent the total number of images and categories, respectively.

\subsection{Few-Shot Attribution Adaptation} \label{sec:adaption}
The registered AI-generated image database contains a limited number of samples for each image generator, including new and unseen ones. The proposed few-shot attribution adaptation leverages this database to efficiently adapt the pretrained model to these unseen generators. 

We first define the image attribution loss. We do not use the commonly applied cross-entropy loss, as it may disrupt the feature representations learned during pretraining. Cross-entropy focuses solely on maximizing classification accuracy and does not explicitly preserve the structure of the feature space. As a result, previously learned discriminative features may be altered, even when fine-tuning with only a few samples. Instead, we incorporate the center loss~\cite{wen2016discriminative} to encourage samples of the same class to cluster around their corresponding class centers. Mathematically, the image attribution loss $\mathcal{L}_A$ is given by:
\begin{equation}
   {\mathcal{L}_A} = \sum\limits_{i = 1}^m {\left\| {{x_i} - {c_{{y_i}}}} \right\|_2^2},
\end{equation}
where $x_i$ denotes the learned feature of the $i$-th sample, $y_i$ is its corresponding attribution category label, and $c_{y_i}$ represents the center of the $y_i$-th category. During training, the center $c_{j}$ is updated within each mini-batch as follows:
\begin{equation}
c_{j}^{t+1} = c_{j}^{t} - \alpha \cdot \frac{\sum_{i=1}^{m} \delta (y_i = j) \cdot  (c_{j}^{t} - x_i)}{1 + \sum_{i=1}^{m} \delta (y_i = j)},
\end{equation}
where $\alpha$ is the learning rate, and $\delta(\cdot)$ is the indicator function that equals one if the condition is true and zero otherwise. The center loss is considered as a regularization which encourages intra-class compactness, constraining the drift of learned features and helping preserve the structure of the pretrained feature space. 

We employ a two-stage paradigm for image attribution, in which Deepfake detection is conducted first, followed by assigning images to their respective generators. Thus, the Deepfake detection loss is also defined. To enable the model to better distinguish the feature discrepancies between real and fake images, we adopt a real-prototype-based contrastive loss, which pulls the features of real images closer to the real prototype while pushing the features of AI-generated images away from it. Formally, the real-prototype-based contrastive loss is defined as:
\begin{align}
    {\mathcal{L}_{D}} =  -& \frac{1}{{{N_r}}}\sum\limits_{i = 1}^{{N_r}} {\log \sigma \big( {\frac{{{\rm{sim}}({x_i^r,{p_r}})}}{\tau }} \big)} \notag \\ 
    -& \frac{1}{{{N_f}}}\sum\limits_{j = 1}^{{N_f}} {\log \Big( {1 - \sigma \big( {\frac{{{\rm{sim}}( {x_i^f,{p_r}})}}{\tau }} \big)} \Big)} ,
\end{align}
where $N_r$ and $N_f$ denote the numbers of real and AI-generated images, respectively; $x_i^r$ and $x_i^f$ represent the learned features of real and AI-generated images, respectively; $p_r$ is the prototype of the real class, which is the averaged feature of all images on the ImageNet; $\sigma(\cdot)$ is the sigmoid function; $\tau$ is a temperature parameter; and $\mathrm{sim}(\cdot, \cdot)$ denotes the cosine similarity. Note that we also avoid using the cross-entropy loss for Deepfake detection in order to preserve the structure of the pretrained feature space.

The final few-shot attribution adaptation loss is: 
\begin{equation}
 \mathcal{L} = {\mathcal{L}_A} + \lambda \mathcal{L}_{D},
\label{eq8}
\end{equation}
where $\lambda$ is the weight parameter.

\section{Experiments}

\subsection{Experimental Setup}

\textbf{Dataset:} 
This study evaluates image attribution on two large-scale benchmarks for AI-generated image detection, namely GenImage~\cite{zhu2023genimage} and WildFake~\cite{hong2025wildfake}.
The GenImage dataset comprises 1,331,167 real images and 1,350,000 synthetic images. The real images are derived from ImageNet, while the synthetic images are generated by eight representative diffusion and GAN models, including Midjourney~\cite{midjourney2022-note}, Stable Diffusion (v1.4/1.5)~\cite{rombach2022high}, ADM~\cite{dhariwal2021diffusion}, GLIDE~\cite{nichol2021glide}, Wukong~\cite{wukong2022}, VQDM~\cite{gu2022vector}, and BigGAN~\cite{DBLP:conf/iclr/BrockDS19}.
The WildFake dataset contains 1,013,446 real images and 2,557,278 synthetic images. The real images are collected from public sources consistent with the training distributions of mainstream generative models, such as COCO~\cite{lin2014microsoft} and ImageNet~\cite{deng2009imagenet}. The synthetic images include those generated by the authors using GANs, diffusion models, and other generation mechanisms, as well as those gathered from open platforms such as Civitai~\cite{Civitai} and Midjourney~\cite{midjourney2022-note}.
The dataset specifies five evaluation levels, and we focus on the cross-generator and cross-architecture settings to assess attribution performance across different levels of granularity, ranging from coarse to fine.

\noindent\textbf{Evaluation Metrics:}
For Deepfake detection, accuracy serves as the evaluation metric for this binary classification task. 
For image attribution, Rank 1 and mean Average Precision (mAP) are reported for evaluation. Rank-1 is the proportion of queries whose top prediction matches the ground truth, while mAP is the mean of Average Precision scores across all queries.

\begin{table*}[th]
\begin{adjustbox}{max width=\textwidth}
\begin{tabular}{c|l|cc|cc|cc|cc|cc|cc|cc|cc|cc|cc}
\toprule
\multirow{2}{*}{Shot} & \multirow{2}{*}{Method} & \multicolumn{2}{c|}{Real} & \multicolumn{2}{c|}{Big} & \multicolumn{2}{c|}{Mid} & \multicolumn{2}{c|}{Wuk} & \multicolumn{2}{c|}{SDV4} & \multicolumn{2}{c|}{SDV5} & \multicolumn{2}{c|}{ADM} & \multicolumn{2}{c|}{GLI} & \multicolumn{2}{c|}{VQ} & \multicolumn{2}{c}{Avg} \\
&  & Rank-1 & mAP & Rank-1 & mAP & Rank-1 & mAP & Rank-1 & mAP & Rank-1 & mAP & Rank-1 & mAP & Rank-1 & mAP & Rank-1 & mAP & Rank-1 & mAP & Rank-1 & mAP \\ \midrule  
\multirow{5}{*}{\rotatebox{90}{1-shot}}    
&ResNet~\cite{he2016deep} & 2.5 & 21.5 & 19.6 & 40.6 & 35.7 & 53.2 & \textbf{46.7} & \textbf{68.8} & 28.6 & 54.3 & 9.5 & 32.7 & 2.0 & 17.6 & 0.5 & 18.6 & 11.1 & 30.1 & 17.4 & 37.5 \\
&DIRE~\cite{wang2023dire} & 13.1 & \textbf{38.8} & 11.6 & 26.8 & 16.1 & 28.0 & 14.6 & 41.7 & 13.6 & 29.3 & 15.1 & 33.9 & 7.0 & 28.0 & 34.2 & 59.8 & 4.0 & 26.8 & 14.3 & 34.8 \\
&ESSP~\cite{chen2024single} & 6.0 & 23.4 & 26.1 & 45.0 & 8.5 & 19.1 & 18.1 & 39.1 & 22.6 & 52.4 & \textbf{22.6} & \textbf{40.7} & 9.0 & 28.2 & 36.2 & 53.6 & 4.0 & 22.1 & 17.0 & 36.0 \\
&Ours & \textbf{21.5} & 22.7 & \textbf{97.0} & \textbf{88.3} & \textbf{74.4} & \textbf{91.6} & 30.2 & 54.3 & \textbf{32.2} & \textbf{63.2} & 1.5 & 31.0 & \textbf{23.6} & \textbf{53.7} & \textbf{40.2} & \textbf{63.6} & \textbf{52.8} & \textbf{75.4} & \textbf{40.4} & \textbf{61.5}\\ \midrule  
\multirow{5}{*}{\rotatebox{90}{5-shot}}    & ResNet~\cite{he2016deep} & 25.1 & 27.5 & 10.1 & 20.4 & 32.2 & 25.2 & 30.7 & 32.3 & 17.6 & 22.6 & 27.6 & 33.4 & 6.5 & 19.8 & 15.6 & 24.8 & 9.0 & 19.5 & 19.4 & 25.0 \\
                      & DIRE~\cite{wang2023dire} & 30.2 & 34.1 & 13.1 & 18.5 & 36.7 & 38.5 & 20.1 & 27.7 & \textbf{26.1} & 25.5 & 8.0 & 20.0 & 16.6 & 21.9 & 14.6 & 21.4 & 3.0 & 15.4 & 18.7 & 24.8 \\
                      & ESSP~\cite{chen2024single} & 16.1 & 22.6 & 14.6 & 17.6 & 30.2 & 30.3 & 20.6 & 24.9 & 11.6 & 21.5 & 10.6 & 23.2 & 11.1 & 20.4 & \textbf{32.7} & 31.8 & 10.1 & 21.1 & 17.5 & 23.7 \\
                      & Ours          & \textbf{76.9} & \textbf{54.5} & \textbf{98.5} & \textbf{98.6} & \textbf{73.9} & \textbf{57.3} & \textbf{32.2} & \textbf{38.4} & 23.6 & \textbf{47.3} & \textbf{43.7} & \textbf{46.1} & \textbf{36.7} & \textbf{56.7} & 32.2 & \textbf{47.1} & \textbf{69.3} & \textbf{60.5} & \textbf{54.1} & \textbf{56.3}\\ \midrule 
\multirow{5}{*}{\rotatebox{90}{10-shot}}   & ResNet~\cite{he2016deep} & 16.1 & 19.5 & 10.6 & 15.4 & 56.3 & 30.0 & \textbf{27.6} & 27.7 & 12.1 & 23.6 & 22.6 & 24.2 & 17.1 & 19.8 & 20.1 & 25.4 & 10.1 & 16.2 & 21.4 & 22.4 \\
                      & DIRE~\cite{wang2023dire} & 22.6 & 22.9 & 11.6 & 19.2 & 0.0 & \textbf{70.2} & 26.6 & 27.0 & 22.1 & 27.0 & 20.6 & 29.2 & 18.1 & 21.9 & 22.6 & 23.9 & 10.6 & 17.7 & 17.2 & 28.8 \\
                      & ESSP~\cite{chen2024single} & 17.6 & 22.1 & 13.6 & 18.0 & 49.7 & 29.4 & 27.1 & 27.9 & 19.6 & 22.7 & 17.6 & 22.3 & 16.6 & 21.6 & 27.1 & 25.5 & 13.1 & 17.1 & 22.4 & 23.0 \\
                      & Ours                    & \textbf{83.4} & \textbf{50.8} & \textbf{98.5} & \textbf{97.9} & \textbf{69.3} & 40.4 & 13.1 & \textbf{30.7} & \textbf{23.6} & \textbf{36.1} & \textbf{50.3} & \textbf{46.8} & \textbf{47.2} & \textbf{48.7} & \textbf{55.3} & \textbf{54.7} & \textbf{45.7} & \textbf{58.5} & \textbf{54.0} & \textbf{51.6}   \\ \bottomrule 
\end{tabular}
\end{adjustbox}
\caption{Performance comparison of AI-generated image attribution on the GenImage dataset under the cross-architecture setting with different numbers of shots. The best score for each shot setting is highlighted in bold.}
\label{tab:genimage_fine_attribution}
\end{table*}

\begin{table*}[th]
\begin{adjustbox}{max width=\textwidth}
\begin{tabular}{c|l|cc|cc|cc|cc|cc|cc|cc|cc|cc|cc|cc|cc}
\toprule
\multirow{2}{*}{Shot} & \multirow{2}{*}{Method} & \multicolumn{2}{c|}{Real} & \multicolumn{2}{c|}{VQVAE} & \multicolumn{2}{c|}{COM} & \multicolumn{2}{c|}{DD} & \multicolumn{2}{c|}{VQDM} & \multicolumn{2}{c|}{BigGAN} & \multicolumn{2}{c|}{StyleGAN} & \multicolumn{2}{c|}{StarGAN} & \multicolumn{2}{c|}{DF-GAN} & \multicolumn{2}{c|}{GALIP} & \multicolumn{2}{c|}{GigaGAN} & \multicolumn{2}{c}{Avg}\\
&  & Rank-1 & mAP & Rank-1 & \multicolumn{1}{c|}{mAP} & Rank-1 & \multicolumn{1}{c|}{mAP} & Rank-1 & \multicolumn{1}{c|}{mAP} & Rank-1 & \multicolumn{1}{c|}{mAP} & Rank-1 & \multicolumn{1}{c|}{mAP} & Rank-1 & \multicolumn{1}{c|}{mAP} & Rank-1 & \multicolumn{1}{c|}{mAP} & Rank-1 & \multicolumn{1}{c|}{mAP} & Rank-1 & \multicolumn{1}{c|}{mAP} & Rank-1 & \multicolumn{1}{c|}{mAP} & Rank-1 & \multicolumn{1}{c}{mAP}       \\ \midrule
\multirow{5}{*}{\rotatebox{90}{1-shot}}  & ResNet~\cite{he2016deep}  & 23.1 & 40.0 & 36.7 & 60.1 & \textbf{31.7} & \textbf{45.5} & 2.0 & 14.3 & 0.5 & 16.9 & 12.6 & 33.1 & 5.0 & 36.9 & 19.1 & 45.8 & 10.1 & 21.2 & 45.7 & 55.6 & 5.0 & 33.5 & 17.4 & 36.6 \\
                       & DIRE~\cite{wang2023dire}  & 9.0 & 26.5 & 20.6 & 37.2 & 4.5 & 23.4 & \textbf{31.2} & \textbf{57.9} & 9.0 & 24.1 & 4.5 & 16.6 & 10.1 & 27.5 & \textbf{32.2} & 55.8 & 17.1 & 42.2 & 47.7 & 58.7 & 1.5 & 15.9 & 17.0 & 35.1 \\
                       & ESSP~\cite{chen2024single}   & 27.6 & 47.7 & 33.7 & 50.3 & 9.5 & 24.5 & 6.0 & 31.5 & 30.2 & 41.2 & 7.5 & 33.6 & 13.1 & 34.1 & 22.6 & 46.5 & 10.1 & 23.1 & 45.7 & 53.6 & 15.1 & 39.4 & 20.1 & 38.7\\
                       & Ours  & \textbf{68.8} & \textbf{79.4} & \textbf{64.8} & \textbf{81.7} & 11.6 & 32.3 & 15.1 & 50.2 & \textbf{60.8} & \textbf{74.6} & \textbf{100.0} & \textbf{100.0} & \textbf{35.7} & \textbf{64.6} & 29.6 & \textbf{52.0} & \textbf{43.7} & \textbf{71.9} & \textbf{56.8} & \textbf{72.3} & \textbf{69.8} & \textbf{81.5} & \textbf{50.6} & \textbf{69.1} \\ 
                       \midrule
\multirow{5}{*}{\rotatebox{90}{5-shot}}      & ResNet~\cite{he2016deep}   & 2.5 & 12.9 & 48.2 & 41.8 & 0.0 & 10.0 & 21.6 & 31.8 & 12.6 & 19.6 & 21.1 & 24.4 & 26.6 & 25.9 & 37.7 & 45.3 & 22.6 & 25.9 & 47.2 & 55.1 & 1.0 & 9.6 & 21.9 & 27.5 \\
                       & DIRE~\cite{wang2023dire}  & 9.5 & 17.1 & 32.7 & 32.2 & 11.6 & 19.3 & 27.6 & 24.2 & 5.0 & 12.5 & 13.6 & 20.7 & \textbf{36.2} & 27.1 & 36.7 & 35.5 & \textbf{27.6} & 24.8 & 48.2 & 54.8 & 30.7 & 26.6 & 25.4 & 26.8 \\
                       & ESSP~\cite{chen2024single}   & 19.1 & 18.9 & 47.7 & 49.4 & 22.1 & 27.8 & 10.1 & 20.8 & 16.6 & 22.2 & 22.6 & 23.7 & 28.6 & 26.8 & 25.6 & 27.4 & \textbf{27.6} & 23.0 & 47.2 & 50.0 & 26.1 & 26.7 & 26.7 & 28.8 \\
                       & Ours & \textbf{69.3} & \textbf{71.5} & \textbf{50.3} & \textbf{65.9} & \textbf{30.2} & \textbf{31.6} & \textbf{44.7} & \textbf{53.8} & \textbf{43.2} & \textbf{54.8} & \textbf{99.5} & \textbf{86.6} & 25.6 & \textbf{34.2} & \textbf{51.8} & \textbf{51.0} & 23.6 & \textbf{48.1} & \textbf{84.4} & \textbf{66.1} & \textbf{86.9} & \textbf{74.6} & \textbf{55.4} & \textbf{58.0} \\
                       \midrule
\multirow{5}{*}{\rotatebox{90}{10-shot}} & ResNet~\cite{he2016deep}  & 25.1 & 26.0 & \textbf{57.8} & 35.9 & 30.2 & 21.9 & 13.1 & 15.1 & 18.6 & 15.6 & 25.6 & 22.4 & 30.7 & 26.1 & 39.7 & 28.4 & 17.6 & 15.3 & 49.2 & 46.8 & 21.1 & 20.7 & 29.9 & 24.9 \\
                       & DIRE~\cite{wang2023dire}   & 31.7 & 20.4 & 56.3 & 43.8 & 22.6 & 22.5 & 19.1 & 18.4 & 28.6 & 20.5 & 10.6 & 14.4 & \textbf{51.8} & 30.9 & 26.1 & 29.4 & 19.1 & 21.1 & 51.8 & 57.5 & 28.6 & 22.4 & 31.5 & 27.4 \\
                       & ESSP~\cite{chen2024single}   & 25.6 & 25.3 & 28.6 & 28.9 & 20.6 & 18.0 & 20.6 & 17.0 & 15.6 & 18.3 & 27.6 & 18.8 & 39.7 & 26.5 & 40.7 & 34.3 & 19.1 & 19.0 & 50.3 & 60.7 & 23.6 & 20.2 & 28.4 & 26.1 \\
                       & Ours & \textbf{67.8} & \textbf{63.0} & 46.7 & \textbf{52.1} & \textbf{54.3} & \textbf{44.3} & \textbf{62.8} & \textbf{45.1} & \textbf{52.8} & \textbf{60.6} & \textbf{97.5} & \textbf{99.0} & 39.2 & \textbf{34.4} & \textbf{55.3} & \textbf{54.3} & \textbf{50.8} & \textbf{53.6} & \textbf{83.4} & \textbf{69.2} & \textbf{74.4} & \textbf{52.6} & \textbf{62.3} & \textbf{57.1} \\
                       \midrule
\end{tabular}
\end{adjustbox}
\caption{Performance comparison of AI-generated image attribution on the WildFake dataset under the cross-architecture setting with different numbers of shots. The best score for each shot setting is highlighted in bold.}
\label{tab:wildfake_fine_attribution}
\end{table*}

\begin{table}[th]
\begin{adjustbox}{max width=0.5\textwidth}
\begin{tabular}{c|l|cc|cc|cc|cc|cc}
\toprule
\multirow{2}{*}{Shot} & \multirow{2}{*}{Method} & \multicolumn{2}{c|}{Real} & \multicolumn{2}{c|}{GAN-based} & \multicolumn{2}{c|}{Midjourney} & \multicolumn{2}{c|}{Diffusion-based} & \multicolumn{2}{c}{Avg} \\
                      &                         & Rank-1         & mAP         & Rank-1         & mAP        & Rank-1         & mAP        & Rank-1         & mAP         & Rank-1        & mAP        \\ \midrule
\multirow{5}{*}{\rotatebox{90}{1-shot}}    & ResNet~\cite{he2016deep} & 44.7 & 71.3 & 6.0 & 34.5 & \textbf{79.9} & \textbf{89.0} & 19.1 & 46.9 & 37.4 & 60.4 \\
                      & DIRE~\cite{wang2023dire} & 27.6 & 55.5 & 18.1 & 48.1 & 26.1 & 49.0 & \textbf{55.3} & \textbf{77.2} & 31.8 & 57.5 \\
                      & ESSP~\cite{chen2024single} & 54.8 & 74.5 & 51.3 & 71.2 & 31.7 & 62.0 & 5.5 & 32.4 & 35.8 & 60.0 \\
                      & Ours    & \textbf{88.4} & \textbf{93.0} & \textbf{96.0} & \textbf{97.9} & 74.9 & 87.4 & 50.8 & 72.6 & \textbf{77.5} & \textbf{87.7}\\ 
                      \midrule
\multirow{5}{*}{\rotatebox{90}{5-shot}}    & ResNet~\cite{he2016deep} & 27.1 & 40.9 & 25.6 & 37.0 & 68.3 & 48.8 & 36.7 & 50.3 & 39.4 & 44.3\\
                      & DIRE~\cite{wang2023dire} & 36.2 & 46.3 & 44.7 & 55.5 & 53.3 & 53.0 & 35.2 & 46.3 & 42.3 & 50.3 \\
                      & ESSP~\cite{chen2024single} & 24.1 & 35.6 & 51.3 & 51.0 & 49.2 & 45.0 & 39.7 & 57.1 & 41.1 & 47.2 \\
                      & Ours     & \textbf{79.4} & \textbf{81.2} & \textbf{98.0} & \textbf{99.1} & \textbf{84.9} & \textbf{78.2} & \textbf{81.9} & \textbf{63.9} & \textbf{86.1} & \textbf{80.6} \\ 
                      \midrule
\multirow{5}{*}{\rotatebox{90}{10-shot}}   & ResNet~\cite{he2016deep} & 48.7 & 47.5 & 43.2 & 41.6 & 66.8 & 51.0 & 18.1 & 34.6 & 44.2 & 43.7 \\
                      & DIRE~\cite{wang2023dire} & 35.7 & 47.9 & 32.2 & 35.4 & 65.3 & 55.1 & 22.6 & 31.2 & 38.9 & 42.4 \\
                      & ESSP~\cite{chen2024single} & 34.7 & 38.8 & 32.2 & 40.7 & 76.9 & 58.8 & 28.1 & 36.4 & 43.0 & 43.7 \\
                      & Ours    & \textbf{89.4} & \textbf{80.4} & \textbf{99.0} & \textbf{98.8} & \textbf{94.0} & \textbf{63.6} & \textbf{72.4} & \textbf{56.2} & \textbf{88.7} & \textbf{74.7}\\ 
                      \bottomrule
\end{tabular}
\end{adjustbox}
\caption{Performance comparison of AI-generated image attribution on the GenImage dataset under the cross-generator setting with different numbers of shots. The best score for each shot setting is highlighted in bold.}
\label{tab:genimage_coarse_attribution}
\end{table}

\noindent\textbf{Implementation Details:}
Following the train-test split protocols of GenImage and WildFake, we construct a registered database by randomly selecting 1, 5, and 10 synthetic images per generator, respectively, from the training set.
All images in the test set are then used as queries to evaluate the performance.
The temperature parameter $\tau$ and the weight parameter $\lambda$ are set to 0.1 and 0.9, respectively, for both datasets.
The pretrained model is adapted to the image database using a batch size of 32 and an initial learning rate of $1\times10^{-4}$, trained for a total of 100 epochs.
All experiments are conducted on an Ubuntu 22.04 system with an RTX 4090 GPU, implemented using PyTorch 2.0.1.

\subsection{Evaluation of AI-generated Image Attribution}
For performance comparison, we construct three baselines, including ResNet-50~\cite{he2016deep} and two models specifically tailored for AI-generated image detection, DIRE~\cite{wang2023dire} and ESSP~\cite{chen2024single}. All these attribution extractors are pre-trained on RGB images and employed to extract features from both the query and registered database images, followed by detection and attribution based on cosine similarity.

\noindent\textbf{Results on GenImage:}
For the GenImage dataset, we first perform image attribution across eight different generative architectures, with Rank-1 and mAP reported in Table~\ref{tab:genimage_fine_attribution}. 
It can be observed that the random guess probability is 11.1\%, whereas our method achieves a Rank-1 exceeding 50\%.
Compared with other attribution extractors, LIDA outperforms ResNet, DIRE, and ESSP in the 10-shot setting in Rank-1 by 32.6\%, 36.8\%, and 31.6\%, respectively, demonstrating that low-bit generative fingerprints effectively capture the noise structures characteristic of different generative architectures.
As the number of shots per generative architecture increases from 1 to 5 and 10, the Rank-1 for our method improves by 4.8\% and 11.7\%, respectively. This suggests that accuracy is positively correlated with the size of the database, as a larger number of reference samples provides richer and more reliable class information, leading to more stable and accurate attribution by the model.
Note that the trend of mAP is opposite to that of Rank-1, as having more retrieval samples makes it more challenging to maintain high-quality ranking. Nevertheless, our method consistently outperforms other attribution methods across different shot settings.

\noindent\textbf{Results on WildFake:}
We observe that some models share similar noise patterns due to their underlying architectures, which makes it challenging to differentiate among them. Therefore, we merge DDPM~\cite{ddpm}, DDIM~\cite{ddim}, and ADM~\cite{adm} into the DD subset, and DALL·E~\cite{dalle}, Imagen~\cite{imagen}, Stable Diffusion~\cite{stable}, and Midjourney~\cite{midjourney2022-note} into the COM subset.
As a result, a total of 10 different generative images need to be attributed.
All experimental results exceed the random guess probability of 9.1\%. Under the 10-shot setting, our method achieves a Rank-1 accuracy of 62.3\%, surpassing ResNet, DIRE, and ESSP by substantial margins of 32.4\%, 30.8\%, and 33.9\%, respectively. Across all subsets, our method achieves the highest attribution performance on BigGAN, attaining a Rank-1 of 100\% in the 1-shot setting, greatly outperforming other attribution extractors. All these results demonstrate the effectiveness of the low-bit generative fingerprints we constructed.

\noindent\textbf{Generator-Level Image Attribution:}
We then combine the six subsets, excluding the BigGAN and Midjourney subsets, to form a diffusion-based set and evaluate performance at the generator level, as shown in Table~\ref{tab:genimage_coarse_attribution}. Without the need to further differentiate between specific diffusion models, only distinguishing between different generative paradigms becomes simpler, resulting in improvements of more than 30\% in Rank-1 and 20\% in mAP, respectively.
It can also be observed that the low-bit generative fingerprint is particularly effective for GAN-based methods, as it boosts the mAP from 41.6\% with ResNet to 98.8\% with our method.

\begin{table}[t]
\begin{adjustbox}{max width=0.5\textwidth}
\begin{tabular}{c|l|ccccccccc}
\toprule
Shot &  Method & Big & Mid & WuK & SDV4 & SDV5 & ADM & GLI & VQ & Avg \\ \midrule
\multirow{4}{*}{\rotatebox{90}{1-shot}}    & ResNet~\cite{he2016deep} & 53.8 & 48.7 & 56.5 & 54.3 & 57.5 & 48.5 & 47.0 & 49.0 & 51.9  \\
& DIRE~\cite{wang2023dire} & 52.0 & 3.5  & 52.0 & 54.0 & 51.3 & 53.0 & 52.5 & 51.5 & 46.2 \\
& ESSP~\cite{chen2024single}& 56.8 & 38.7 & 51.0 & 51.5 & 51.3 & 52.7 & 49.0 & 49.5 & 50.1 \\
& Ours & \textbf{86.8}   & \textbf{85.4}   & \textbf{87.2}   & \textbf{85.6}    & \textbf{84.5}    & \textbf{86.5}   & \textbf{89.1}   & \textbf{88.1}  & \textbf{86.5}   \\ \midrule
\multirow{4}{*}{\rotatebox{90}{5-shot}}    & ResNet~\cite{he2016deep} & 55.3 & 57.8 & 55.5 & 52.3 & 53.5 & 52.5 & 56.0 & 56.0 & 54.9   \\
& DIRE~\cite{wang2023dire} & 55.8 & 38.7  & 54.3 & 56.3 & 54.0 & 52.5 & 44.5 & 52.0 & 51.0 \\
& ESSP~\cite{chen2024single}& 53.3 & 55.8 & 49.3 & 52.0 & 51.8 & 53.3 & 52.5 & 49.3 & 52.1 \\
& Ours & \textbf{87.2} & \textbf{84.8} & \textbf{89.4} & \textbf{83.7} & \textbf{85.8} & \textbf{85.7} & \textbf{88.8} & \textbf{88.8}  & \textbf{86.8} \\ \midrule
\multirow{6}{*}{\rotatebox{90}{10-shot}}   & ResNet~\cite{he2016deep} & 52.3 & 58.8 & 55.8 & 55.0 & 60.5 & 54.0 & 60.8 & 60.1 & 57.1 \\
& DIRE~\cite{wang2023dire} & 59.8 & 51.7  & 51.5 & 53.3 & 52.3 & 56.5 & 60.8 & 52.5 & 54.8 \\
& ESSP~\cite{chen2024single}& 54.7 & 58.5 & 53.7 & 58.5 & 52.4 & 54.2 & 51.9 & 57.5 & 55.2 \\
& LARE2$^{\star}$~\cite{luo2024lare} & 72.0 & 62.7 & 79.6 & 79.6 & 79.6 & 63.5 & 80.2 & 76.9 & 72.5 \\
& FSD$^{\star}$~\cite{wu2025fsd} & 82.2 & 80.9  & \textbf{88.8} & 88.8 & \textbf{88.8} & 79.2 & \textbf{97.1} & 76.2 & 84.1 \\
& Ours                    & \textbf{88.1}   & \textbf{89.4}   & 87.4   & \textbf{89.7}    & 85.1    & \textbf{86.1}   & 90.7   & \textbf{90.2}  & \textbf{88.3}  \\ \bottomrule
\end{tabular}
\end{adjustbox}
\caption{Accuracy of AI-generated image detection on the GenImage dataset with different numbers of shots. ($\star$) denotes results taken from \cite{wu2025fsd} and best score for each shot setting is highlighted in bold.}
\label{tab:few_deepfake}
\end{table}

\subsection{Evaluation of AI-generated Image Detection}

\noindent\textbf{Few-Shot Detection:}
For AI-generated image detection, our method consistently achieves the highest average accuracy across different shot settings, as shown in Table~\ref{tab:few_deepfake}. Under the 10-shot setting, it outperforms the state-of-the-art few-shot Deepfake detection method FSD~\cite{wu2025fsd} by 4.2\% on the GenImage dataset. Unlike FSD, which merges WuK, SDV4, and SDV5 into a single subset, our approach preserves the original fine-grained subset partition and still achieves over 85\% accuracy across all subsets. This highlights the forensic effectiveness of our low-bit generative fingerprints and the strength of our adaptation strategy.

\begin{table}[t]
\begin{adjustbox}{max width=0.5\textwidth}
\begin{tabular}{l|ccccccccc}
\toprule
Method & Big & Mid & WuK & SDV4 & SDV5 & ADM & GLI & VQ & Avg \\ \midrule
RIGID~\cite{he2024rigid} & 53.0 & 94.1 & 87.8 & 87.0 & 87.2 & 51.4 & 45.9 & 52.2 & 69.8 \\
AEROBLADE~\cite{ricker2024aeroblade} & 58.3 & 40.2 & 51.4 & 52.6 & 55.1 & 50.7 & 29.4 & 52.8 & 48.8 \\
Manifold~\cite{brokman2025manifold}& 77.6 & 55.5 & 65.4 & 62.0 & 63.0 & 57.3 & 88.3 & 76.9 & 68.2 \\
FSD~\cite{wu2025fsd}& 62.1 & 75.1 & 88.0 & 88.0 & 88.0 & 74.1 & 93.9 & 69.1 & 77.1 \\
Ours & 91.0 & 85.9 & 86.2 & 86.3 & 86.8 & 85.5 & 83.9 & 84.5 & 86.3 \\ 
\bottomrule
\end{tabular}
\end{adjustbox}
\caption{Accuracy of zero-shot AI-generated image detection on the GenImage dataset.}
\label{tab:zs_deepfake}
\end{table}

\noindent\textbf{Zero-Shot Detection:}
We further evaluate the model’s performance under the zero-shot setting, which can be regarded as the lower bound of its detection capability. Specifically, we first create low-bit generative fingerprints from ImageNet to pretrain an adapted ResNet-50, as described in Eq.~(4). This pretrained model is then used to extract features from query images, which are subsequently compared with the mean feature vector of all real images used during pretraining. By manually selecting a classification threshold of 0.85, queries with similarity above this value are considered real, while those below are classified as fake. As shown in Table~\ref{tab:zs_deepfake}, even without any prior knowledge of fake images, our method achieves an accuracy of 86.3\%, surpassing RIGID, AEROBLADE, Manifold, and FSD by 16.5\%, 37.5\%, 18.1\%, and 9.2\%, respectively. 
All of these competitors are specifically designed for zero-shot Deepfake detection. 
The high accuracy achieved by our method under zero-shot settings indicates that the extracted features are sufficiently discriminative.

\subsection{Ablation Studies and Analyses}

\noindent\textbf{Ablation Studies:}
In Table~\ref{tab:ablation}, we discuss the effectiveness of bit-plane-based fingerprints (BF) and three types of loss functions: 
\textbf{(1) Effectiveness of BF}: Comparing results between rows 1 and 2, when using raw images or low-bit generative fingerprints from the registered database as input to a ResNet-50 pretrained on ImageNet, the latter achieves an average mAP that is 10.6\% mAP higher than the former. This demonstrates the effectiveness of BF in capturing generator-specific noise structures.
\textbf{(2) Effectiveness of unsupervised pre-training}: Comparing results between rows 2 and 3, by training the model with BF of real images under supervision of the pretext task training loss $L_P$, the model achieves an additional 1.5\% mAP, highlighting the importance of unsupervised pre-training.
\textbf{(3) Effectiveness of attribution loss $L_A$}: Comparing results between rows 3 and 4, adapting the model to the registered database with only one fake image per generator under supervision of the attribution loss $L_A$ achieves an mAP of 53.3\%, outperforming the model without adaptation by 3.7\%. This indicates that the attribution loss effectively guides sample features to cluster around their corresponding class centers.
\textbf{(4) Effectiveness of Deepfake detection loss $L_D$}: Comparing results between rows 4 and 5, incorporating the Deepfake detection loss $L_D$ further improves overall performance by 8.2\%. This suggests that the real-prototype-based contrastive loss effectively increases the separation between real and fake images in the feature space.

\noindent \textbf{Comparison of Different Losses during Adaptation:} In Sec.~\ref{sec:adaption}, we choose the center loss and the contrastive loss as the attribution loss $L_A$ and the Deepfake detection loss $L_D$, respectively. Table~\ref{tab:loss_diff} presents the results of replacing these losses with the cross-entropy loss.
Replacing either $L_A$ or $L_D$ with cross-entropy results in performance degradations of 1.8\% and 0.8\%, respectively, while substituting both leads to a drop of 3.9\%.

\begin{figure}[t]
    \centering
    \begin{minipage}[t]{0.245\textwidth}
        \centering
        \includegraphics[width=\textwidth,height=0.122\textheight,keepaspectratio=false]{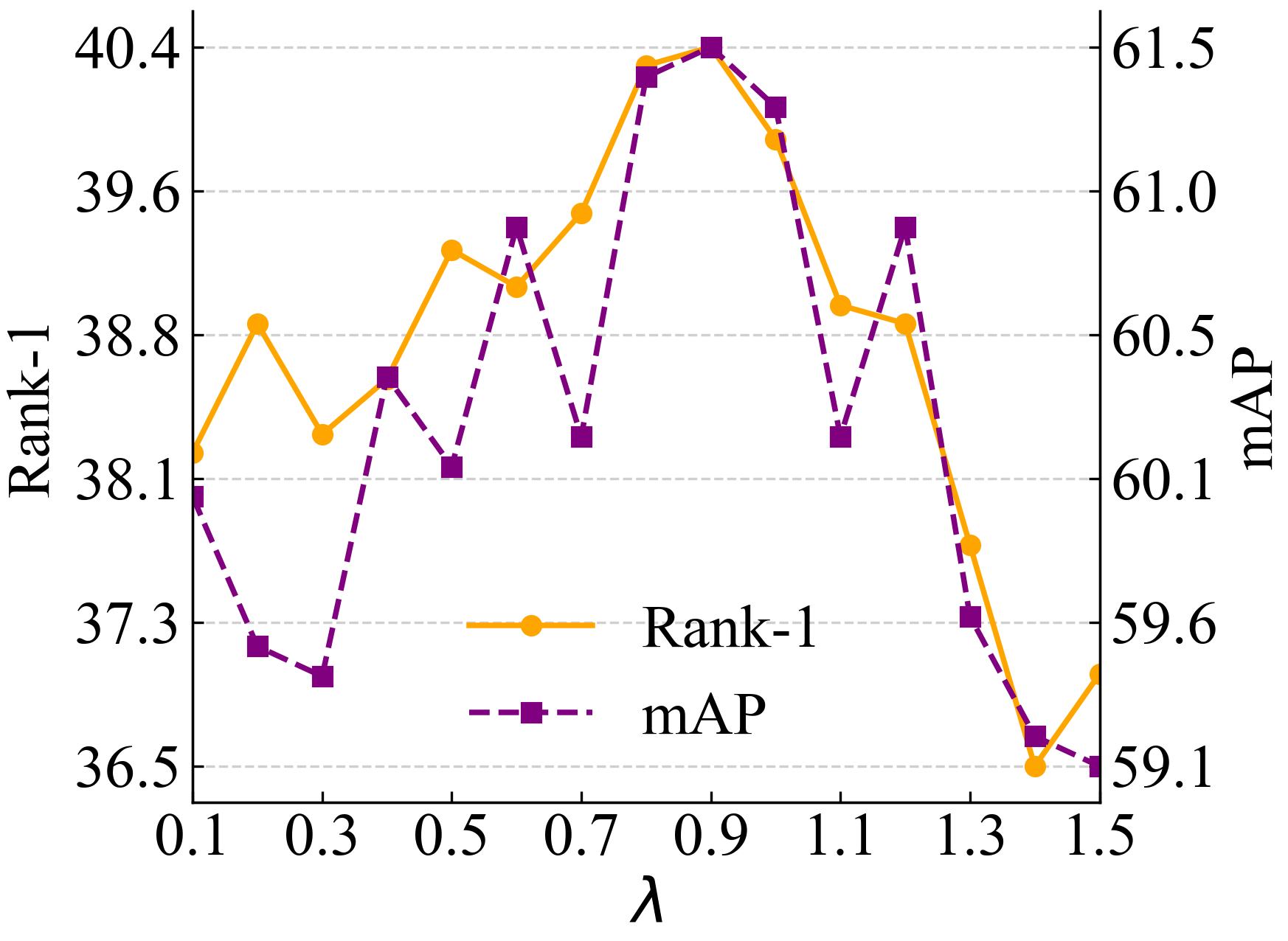}
        \subcaption{}
    \end{minipage}
    \begin{minipage}[t]{0.225\textwidth}
        \centering
        \includegraphics[width=\textwidth]{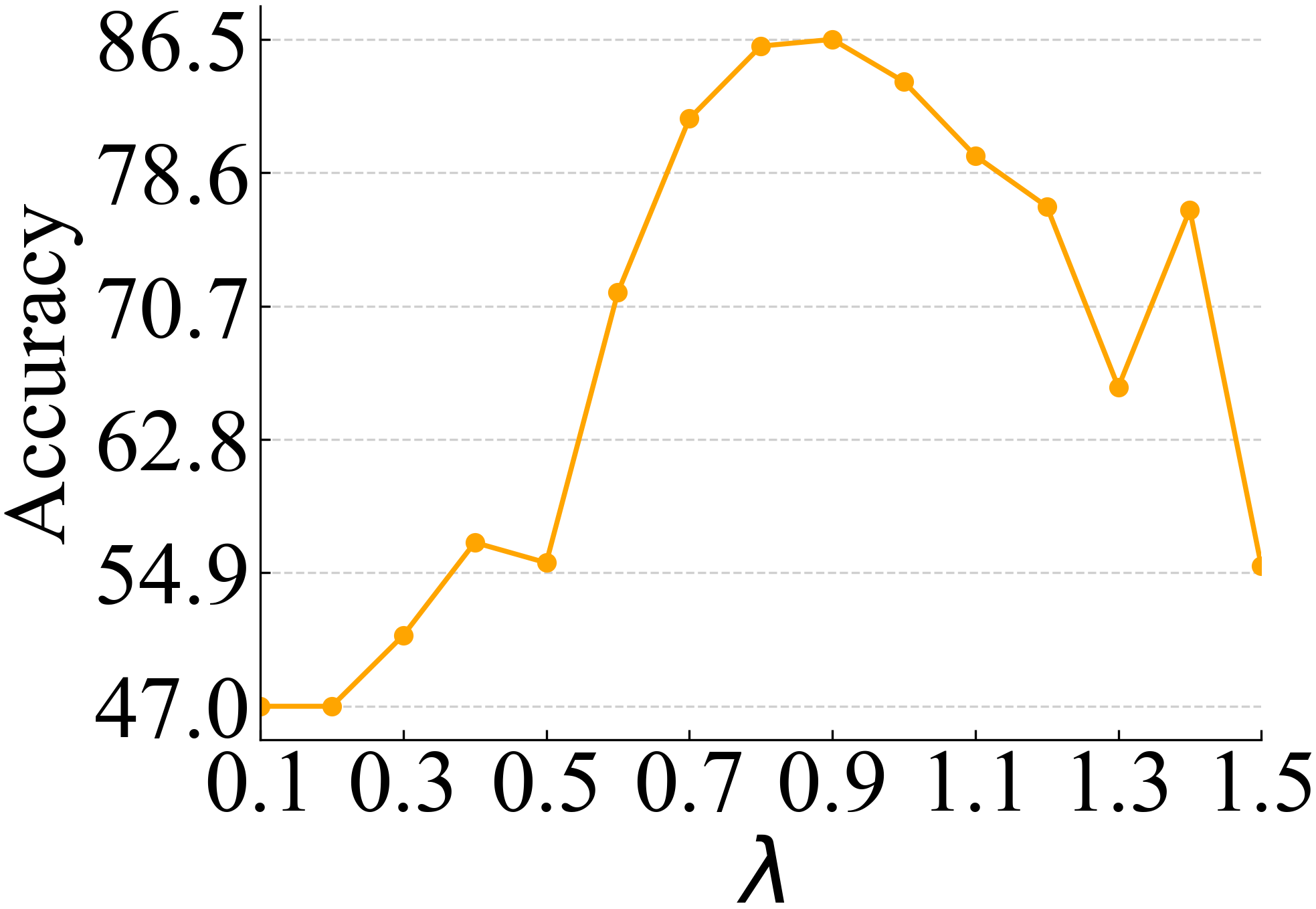}
        \subcaption{}
    \end{minipage}
    \vspace{-0.2cm}
    \caption{\textbf{Impact of loss weight} for (a) AI-generated image attribution and (b) AI-generated image detection.} 
    \label{fig:parameter}
\end{figure}

\begin{figure}[t]
    \centering
    \includegraphics[width=\linewidth]{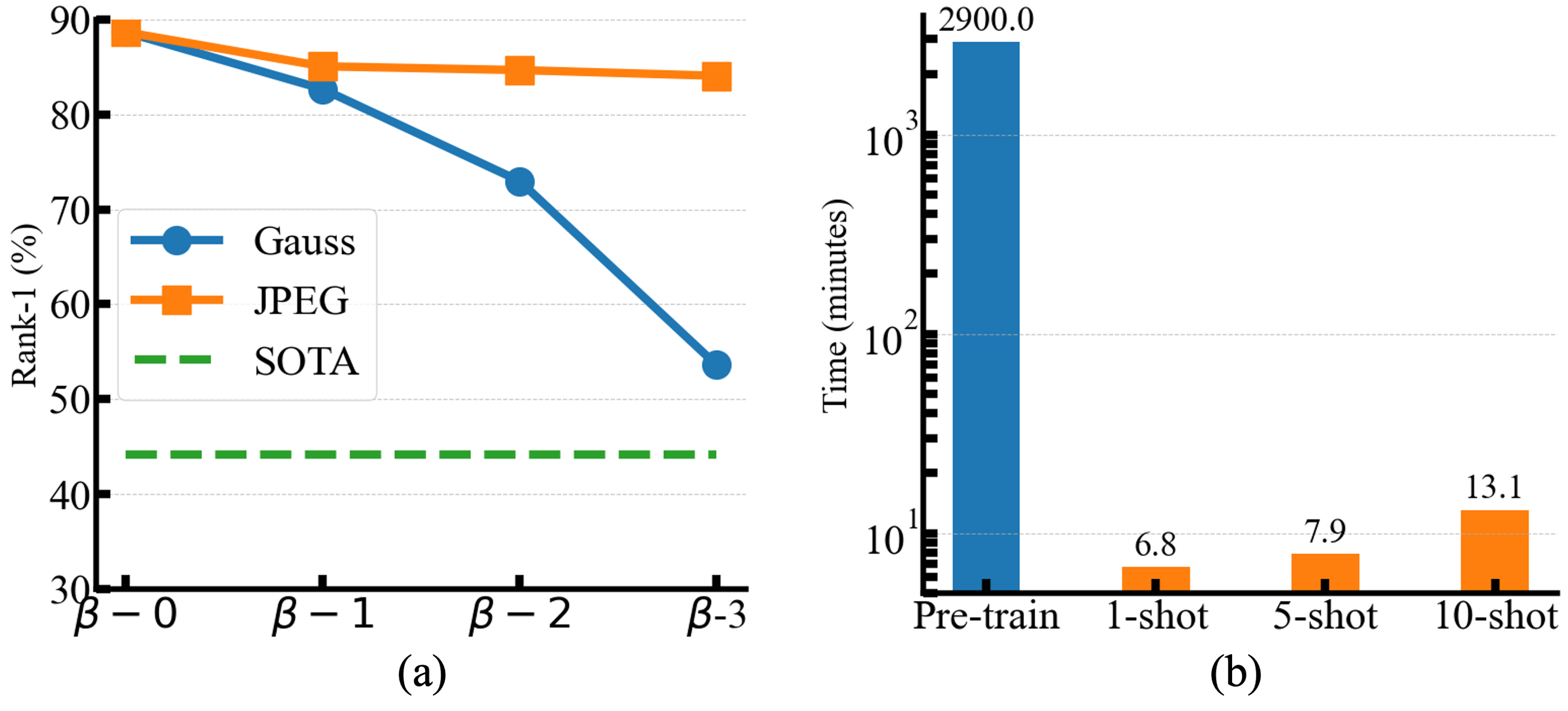}
    \caption{\textbf{Robustness analysis and comparison of adaptation time: }
    (a) Robustness to image degradation under Gaussian blur and JPEG compression, where $\beta$ represents the degree of blurring and image quality. (b) Comparison of the running time for unsupervised pre-training and few-shot attribution adaptation.}
    \label{fig:robustness}
\end{figure}

\noindent\textbf{Impact of Loss Weight $\lambda$:}
In Figure~\ref{fig:parameter}, we analyze the impact of the loss-balancing hyperparameter $\lambda$ in Eq.~(\ref{eq8}). Both Rank-1 and mAP increase as $\lambda$ grows, reaching their peak at $\lambda = 0.9$ (ACC = 40.4\%, mAP = 61.5\%), while accuracy also attains its maximum of 86.5\%. Beyond this value, performance declines, indicating that placing excessive emphasis on separating real and fake samples can hinder the discrimination among different fake generators.

\noindent \textbf{Robustness of Image Degradation:}
To assess the robustness of our forensic method, we apply Gaussian blur ($\theta = 0, 1, 2, 3$) and JPEG compression (quality = 100\%, 95\%, 90\%, 85\%) to the raw images during testing, as shown in Figure~\ref{fig:robustness}(a). A larger value of the horizontal-axis parameter $\beta$ indicates stronger degradation or lower JPEG quality. Our method exhibits strong robustness under varying levels of JPEG compression. 
Gaussian blur directly distorts the distribution of low-bit generative fingerprints. However, even under such degradation, the resulting features still preserve generator-specific noise patterns far more effectively than RGB-based features from unaltered images.

\noindent \textbf{Practical Efficiency Analysis:}
We report the training time for both unsupervised pre-training and attribution adaptation in Figure~\ref{fig:robustness}(b). The running time overhead introduced by our few-shot attribution adaptation is negligible compared to pre-training. When encountering unseen AI-generated images, the proposed retrieval-based attribution paradigm enables accurate attribution through rapid adaptation, overcoming the limitations of conventional classification-based approaches that require full retraining. Since low-bit fingerprint generation relies on efficient binary operations and the attribution encoder is based on ResNet-50, our model operates at millisecond-level inference speed.

\begin{table}[t]
\begin{adjustbox}{max width=0.5\textwidth}
\begin{tabular}{cccc|ccccccccc|c}
\hline
BF & $L_P$ & $L_A$ & $L_D$ &  Real & Big & Mid & Wuk & SDV4 & SDV5 & ADM & GLI & VQ & Avg \\ \hline
\ding{55} & \ding{55} & \ding{55} &\ding{55} &  21.5 & 40.6 & 53.2 & 68.8 & 54.3 & 32.7 & 17.6 & 18.6 & 30.1 & 37.5 \\
\ding{51} & \ding{55} & \ding{55} &\ding{55} & 4.8& 87.9& 22.7& 52.7& 56.8& 33.6& 78.2& 46.6& 49.6& 48.1 \\
\ding{51} & \ding{51} & \ding{55} &\ding{55} & 46.0 & 97.1 & 38.6 & 30.8 & 42.7 & 43.3 & 55.7 & 45.5 & 46.7 & 49.6 \\ 
\ding{51} & \ding{51} & \ding{51} &\ding{55} & 56.1 & 97.5 & 47.5 & 34.7 & 54.7 & 40.4 & 44.6 & 55.4 & 48.5 & 53.3\\
\ding{51} & \ding{51} & \ding{51} &\ding{51} & 22.7 & 88.3 & 91.6 & 54.3 & 63.2 & 31.0 & 53.7 & 63.6 & 75.4 & 61.5 \\ \hline
\end{tabular}
\end{adjustbox}
\caption{Ablation study on the effectiveness of bit-plane-based fingerprints (BF) and different loss functions for one-shot AI-generated image attribution on the GenImage dataset. $L_p$, $L_A$, and $L_D$ denote the pretext task training loss, image attribution loss, and Deepfake detection loss, respectively.}
\label{tab:ablation}
\end{table}

\begin{table}[t]
\begin{adjustbox}{max width=0.5\textwidth}
\begin{tabular}{cc|ccccccccc|c}
\hline
$L_A$ & $L_D$ &  Real & Big & Mid & Wuk & SDV4 & SDV5 & ADM & GLI & VQ & Avg \\ \hline
CE &  CE & 68.8& 94.7& 33.5& 41.5& 32.5& 71.2& 45.6& 50.6& 79.9& 57.6\\
CE & -- & 72.2& 93.4& 57.7& 38.9& 53.3& 64.6& 42.1& 50.3& 64.6& 59.7\\
-- & CE & 47.8& 93.5& 65.2& 67.2& 60.3& 28.3& 76.1& 58.7& 49.3& 60.7\\
-- & --  & 22.7 & 88.3 & 91.6 & 54.3 & 63.2 & 31.0 & 53.7 & 63.6 & 75.4 & 61.5 \\ \hline
\end{tabular}
\end{adjustbox}
\caption{Effects of replacing the losses with cross-entropy in few-shot attribution adaptation. CE denotes the cross-entropy loss and `--' denotes the original loss. }
\label{tab:loss_diff}
\end{table}


\begin{table}[t]
\begin{adjustbox}{max width=\columnwidth}
\begin{tabular}{c|l|cc|cc|cc|cc|cc}
\toprule
\multirow{2}{*}{Shot} & \multirow{2}{*}{Method} & \multicolumn{2}{c|}{Real} & \multicolumn{2}{c|}{Others} & \multicolumn{2}{c|}{DMs} & \multicolumn{2}{c|}{GANs} & \multicolumn{2}{c}{Avg} \\
                      &                         & Rank-1         & mAP         & Rank-1         & mAP        & Rank-1         & mAP        & Rank-1         & mAP         & Rank-1        & mAP        \\ \midrule
\multirow{5}{*}{\rotatebox{90}{1-shot}}    
& ResNet~\cite{he2016deep}  & 1.5 & 29.7 & 21.6 & 53.0 & 30.2 & 55.7 & \textbf{50.8} & \textbf{73.6} & 26.0 & 53.0  \\
& DIRE~\cite{wang2023dire} & 48.7 & 74.3 & \textbf{83.4} & \textbf{91.1} & 5.0 & 34.8 & 3.5 & 30.7 & 35.2 & 57.7 \\
& ESSP~\cite{chen2024single} & 7.0 & 41.0 & 76.4 & 85.9 & 10.6 & 35.2 & 29.1 & 60.9 & 30.8 & 55.7 \\
                      & Ours  & \textbf{80.4} & \textbf{88.4} & 81.4 & 90.7 & \textbf{48.7} & \textbf{62.0} & 18.6 & 56.2 & \textbf{57.3} & \textbf{74.3} \\ 
                      \midrule
\multirow{5}{*}{\rotatebox{90}{5-shot}}    & ResNet~\cite{he2016deep} & 23.6 & 44.6 & \textbf{69.8} & 67.6 & 9.5 & 29.3 & \textbf{50.3} & 47.9 & 38.3 & 47.4 \\
& DIRE~\cite{wang2023dire} & 39.7 & 41.3 & 50.3 & 54.5 & 25.6 & 37.0 & 36.2 & \textbf{50.2} & 37.9 & 45.8 \\
& ESSP~\cite{chen2024single} & 40.2 & 44.9 & 60.3 & 63.9 & 23.6 & \textbf{41.4} & 34.2 & 40.1 & 39.6 & 47.5 \\
& Ours & \textbf{90.5} & \textbf{80.3} & 66.3 & \textbf{86.7} & \textbf{35.2} & 38.2 & 47.2 & 48.8 & \textbf{59.8} & \textbf{63.5}\\ 
\midrule
\multirow{5}{*}{\rotatebox{90}{10-shot}}   & ResNet~\cite{he2016deep}  & 53.3 & 42.1 & 80.4 & 59.1 & 23.6 & 33.4 & 27.1 & 39.0 & 46.1 & 43.4  \\
& DIRE~\cite{wang2023dire} & 44.2 & 46.1 & 58.8 & 47.6 & 28.6 & 36.7 & 27.6 & 36.2 & 39.8 & 41.7 \\
& ESSP~\cite{chen2024single} & 47.2 & 39.4 & 58.3 & 59.9 & 19.1 & 34.9 & 20.6 & 31.8 & 36.3 & 41.5 \\
& Ours    & \textbf{83.9} & \textbf{81.9} & \textbf{86.4} & \textbf{87.4} & \textbf{83.9} & \textbf{54.9} & \textbf{35.7} & \textbf{44.1} & \textbf{72.5} & \textbf{67.1}  \\ 
\bottomrule
\end{tabular}
\end{adjustbox}
\caption{Performance comparison of AI-generated image attribution on the WildFake dataset under the cross-generator setting with different numbers of shots. The best score for each shot setting is highlighted in bold.}
\label{tab:wildfake_coarse_attribution}
\end{table}

\noindent{\textbf{Generator-Level Image Attribution:}}
We further categorize the generative algorithms into three groups based on their model architectures: diffusion-based (DDPM~\cite{ddpm}, DDIM~\cite{ddim}, ADM~\cite{adm}, DALL·E~\cite{dalle}, Stable Diffusion~\cite{stable}, Midjourney~\cite{midjourney2022-note}, Imagen~\cite{imagen}, VQDM~\cite{gu2022vector}), GAN-based (BigGAN~\cite{biggan}, StyleGAN~\cite{style1,style2,style3}, GigaGAN~\cite{giga}, StarGAN~\cite{star1,star2}, DF-GAN~\cite{dfgan}, GALIP~\cite{galip}), and other (VQVAE~\cite{vqvae}), to evaluate cross-generator performance.
As shown in Table~\ref{tab:wildfake_coarse_attribution}, our method achieves the best performance across all evaluation settings.
More specifically, the proposed low-bit generative fingerprints outperform the features extracted by DIRE and ESSP by 25.4\% and 36.3\% in terms of mAP, respectively, indicating their stronger suitability for fake image attribution.
As the number of images in the registered database increases, Rank-1 accuracy improves while the task of ranking similarities across all images becomes progressively more challenging.

\begin{table}[t]
\begin{adjustbox}{max width=0.5\textwidth}
\begin{tabular}{c|l|cccccccccccccccc}
\toprule
Shot & Method & VQV & COM & DD & VQD & Big & Sty & Sta & DF & GA & Gi & Avg \\ \midrule
\multirow{4}{*}{\rotatebox{90}{1-shot}}                         
                                           & ResNet~\cite{he2016deep} &55.5 &39.4 &53.8 &53.0 &56.3 &52.3 &54.0 &36.7 &57.0 &46.7 &50.5 \\
                                           & DIRE~\cite{wang2023dire} &51.3 &49.0 &48.0 &43.9 &42.7 &50.3 &51.5 &51.0 &51.5 &51.0 &49.2 \\
                                           & ESSP~\cite{chen2024single} &52.3 &55.3 &46.2 &51.3 &54.8 &54.3 &48.9 &46.2 &60.3 &38.9 &50.8 \\
                                           & Ours & \textbf{84.4} & \textbf{70.4} & \textbf{85.2} & \textbf{85.2} & \textbf{83.9} & \textbf{85.9} & \textbf{85.2} & \textbf{83.9} & \textbf{84.7} & \textbf{66.8} & \textbf{81.6} \\
                                           \midrule
\multirow{4}{*}{\rotatebox{90}{5-shot}}    & ResNet~\cite{he2016deep} &51.5 &52.5 &50.5 &49.0 &49.0 &52.0 &54.8 &51.8 &55.3 &48.7 &51.5 \\
                                           & DIRE~\cite{wang2023dire} &52.5 &48.7 &53.2 &48.7 &52.7 &51.5 &52.0 &52.7 &53.0 &51.0 &51.6 \\
                                           & ESSP~\cite{chen2024single} &53.0 &49.7 &52.7 &53.8 &56.8 &53.5 &56.3 &52.5 &56.0 &50.0 &53.4 \\
                                           & Ours & \textbf{85.4} & \textbf{82.2} & \textbf{84.9} & \textbf{84.7} & \textbf{85.2} & \textbf{84.7} & \textbf{85.9} & \textbf{85.7} & \textbf{86.4} & \textbf{83.7} & \textbf{84.9} \\
                                           \midrule
\multirow{4}{*}{\rotatebox{90}{10-shot}}   & ResNet~\cite{he2016deep} &59.5 &43.2 &57.3 &56.8 &58.8 &60.8 &61.3 &53.0 &54.5 &58.0 &56.3 \\
                                           & DIRE~\cite{wang2023dire} &59.3 &51.8 &54.5 &52.5 &57.3 &58.0 &58.5 &56.3 &59.8 &50.8 &55.9 \\
                                           & ESSP~\cite{chen2024single} &54.8 &50.5 &52.3 &58.0 &57.9 &57.2 &56.2 &50.8 &55.0 &51.5 &54.4 \\
                                           & Ours & \textbf{90.7} & \textbf{85.2} & \textbf{89.9} & \textbf{91.2} & \textbf{90.2} & \textbf{89.9} & \textbf{90.7} & \textbf{91.7} & \textbf{89.4} & \textbf{87.2} & \textbf{89.6} \\
                                           \bottomrule
\end{tabular}
\end{adjustbox}
\caption{Accuracy of AI-generated image detection on the WildFake dataset with different numbers of shots. The best score for each shot setting is highlighted in bold.}
\label{tab:fw_deepfake2}
\end{table}

\noindent{\textbf{Deepfake Detection:}}
A similar trend is observed for the Deepfake detection task on the WildFake dataset, where our model consistently outperforms all competitors, as shown in Table~\ref{tab:fw_deepfake2}. Specifically, under the 1-shot, 5-shot, and 10-shot settings, our model exceeds ESSP by 30.8\%, 31.5\%, and 35.2\%, respectively, and outperforms DIRE by 32.4\%, 33.3\%, and 33.7\%, respectively.
Notably, all baseline methods remain close to random guessing across generators and shot settings, whereas our method performs substantially better, achieving over 80\% accuracy, which demonstrates its superior cross-architecture generalization.

\begin{figure*}[ht]
    \centering
    \begin{minipage}[ht]{\textwidth}
        \centering
        \includegraphics[width=1\textwidth]{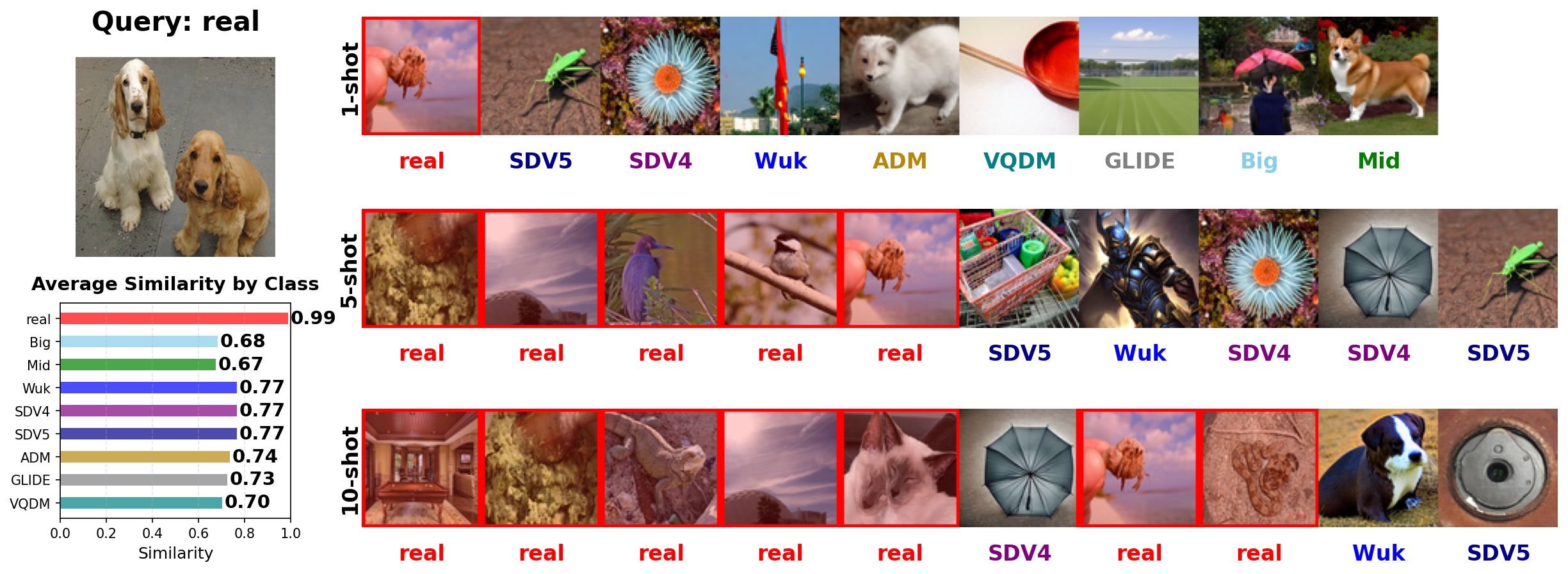}
        \subcaption{Real query image}
    \end{minipage}
    
    \vspace{0cm} 

    \begin{minipage}[ht]{\textwidth}
        \centering
        \includegraphics[width=1\textwidth]{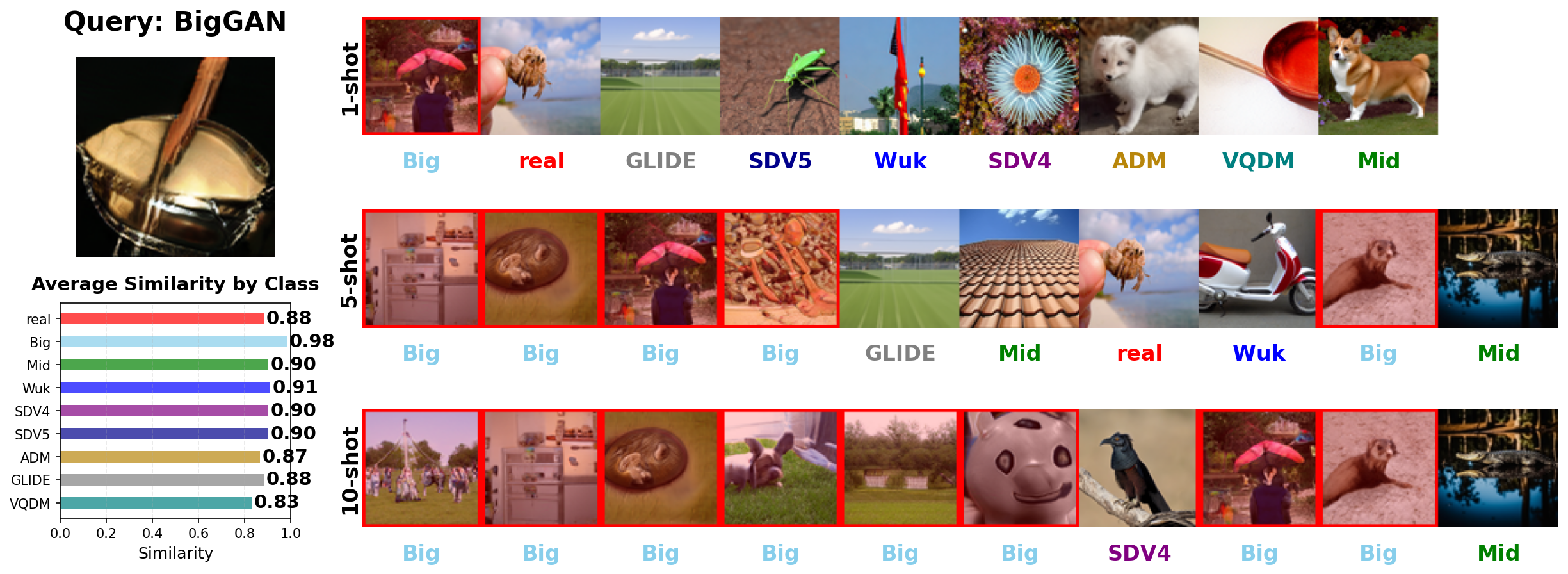}
        \subcaption{Big-GAN generated query image}
    \end{minipage}

    \vspace{0cm} 

    \begin{minipage}[ht]{\textwidth}
        \centering
        \includegraphics[width=1\textwidth]{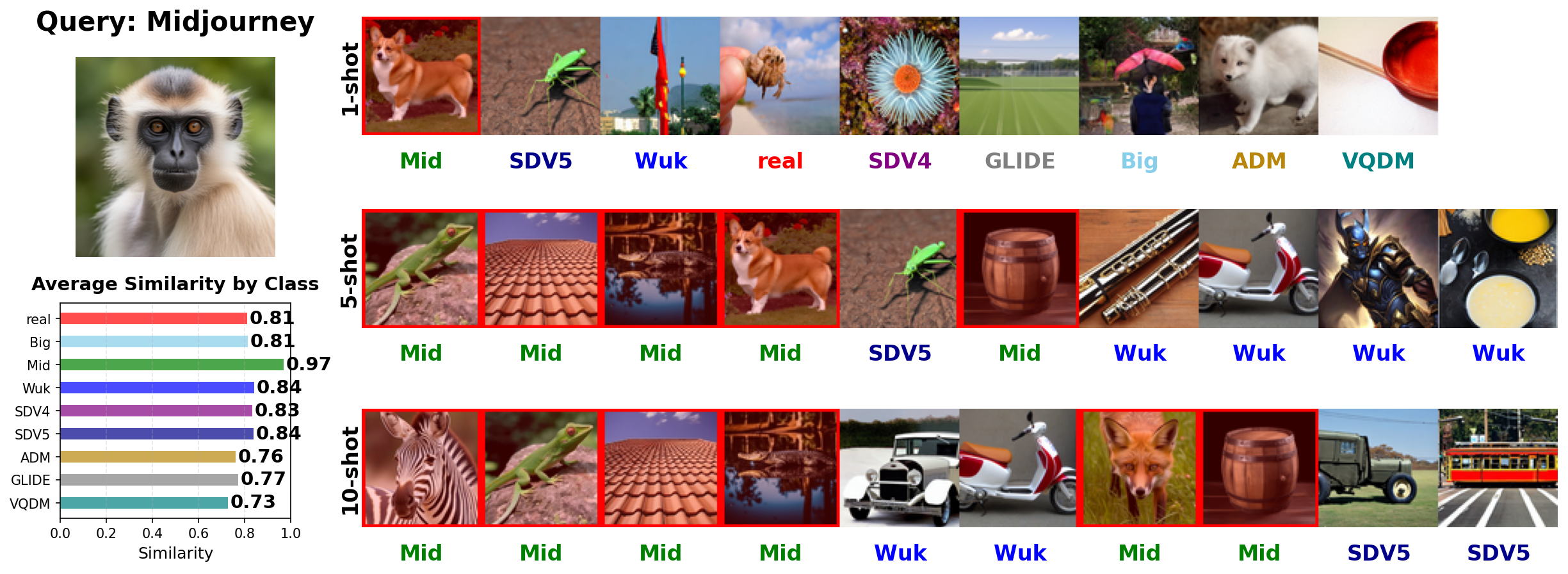}
        \subcaption{Midjourney generated query image}
    \end{minipage}

    \caption{\textbf{Illustration of retrieval-based image attribution.} The top-ranked retrieval results from the database are listed, with the correct result highlighted by a red bounding box. The queries are a real image, a BigGAN-generated~\cite{biggan} image, and a Midjourney-generated~\cite{midjourney2022-note} image, respectively.}
    \label{fig:visual_retrieval_1}
\end{figure*}

\begin{figure*}[ht]
    \centering

    \begin{minipage}[ht]{\textwidth}
        \centering
        \includegraphics[width=1\textwidth]{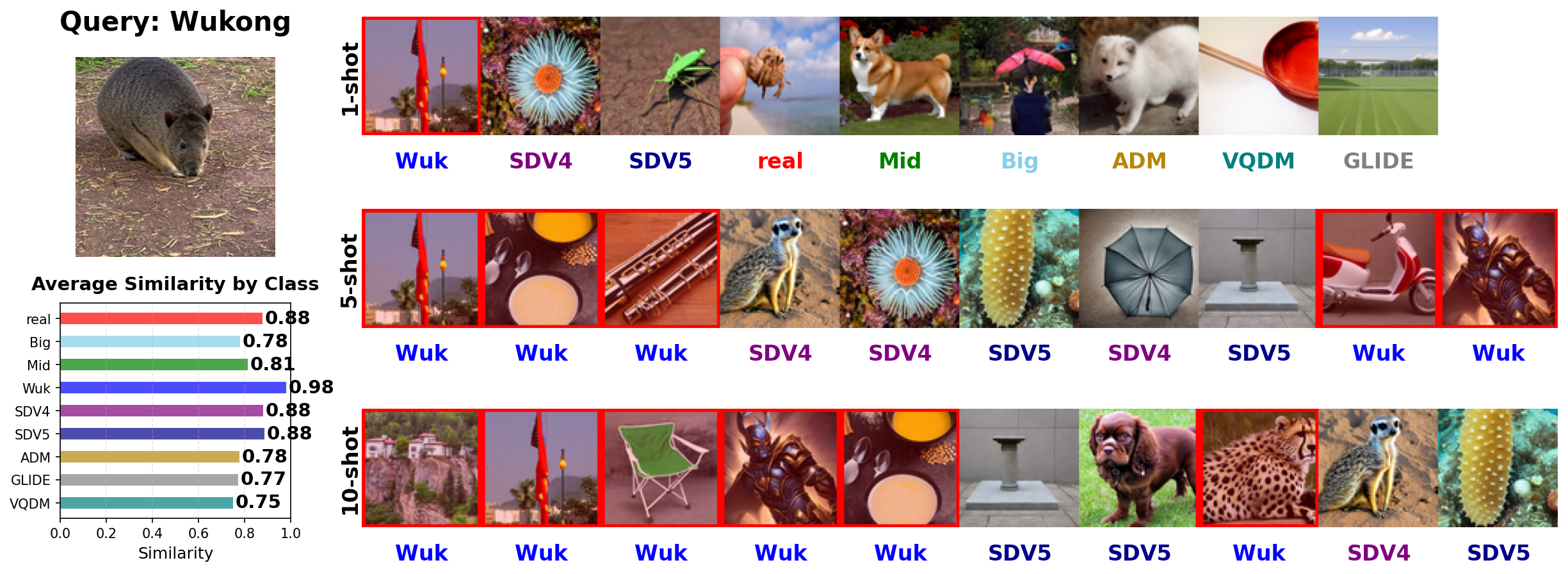}
        \subcaption{Wukong generated query image}
    \end{minipage}

    \begin{minipage}[ht]{\textwidth}
        \centering
        \includegraphics[width=1\textwidth]{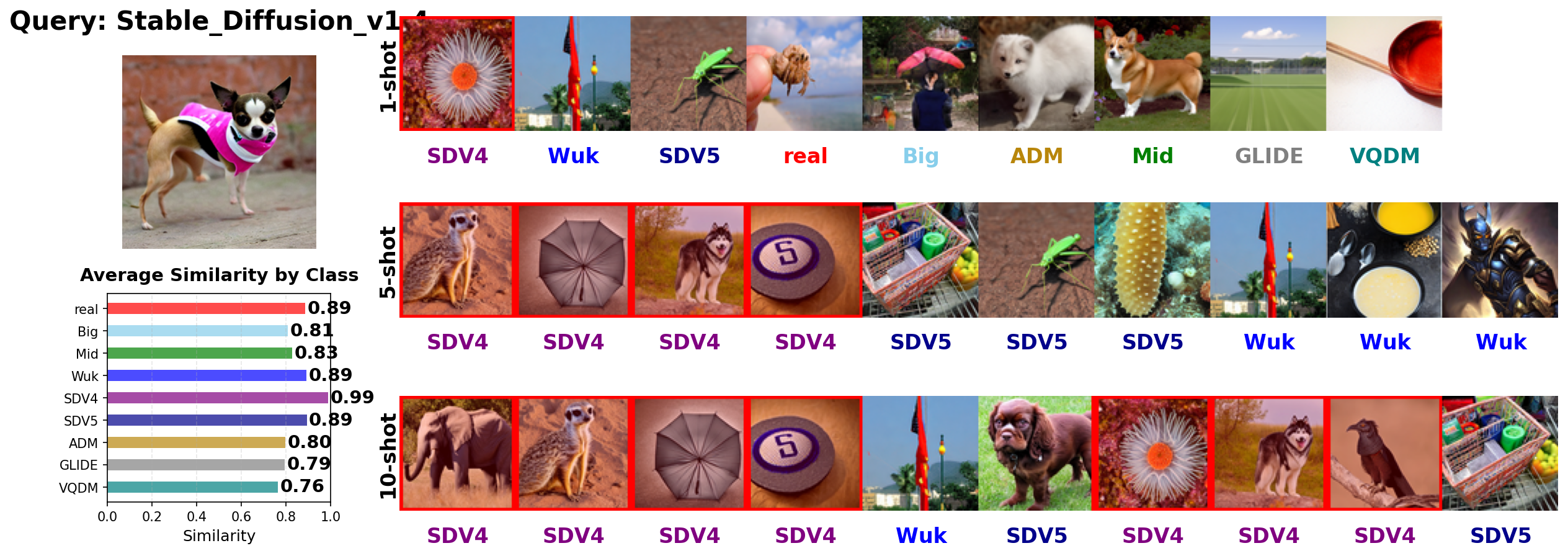}
        \subcaption{Stable Diffusion V1.4 generated query image}
    \end{minipage}

    \begin{minipage}[ht]{\textwidth}
        \centering
        \includegraphics[width=1\textwidth]{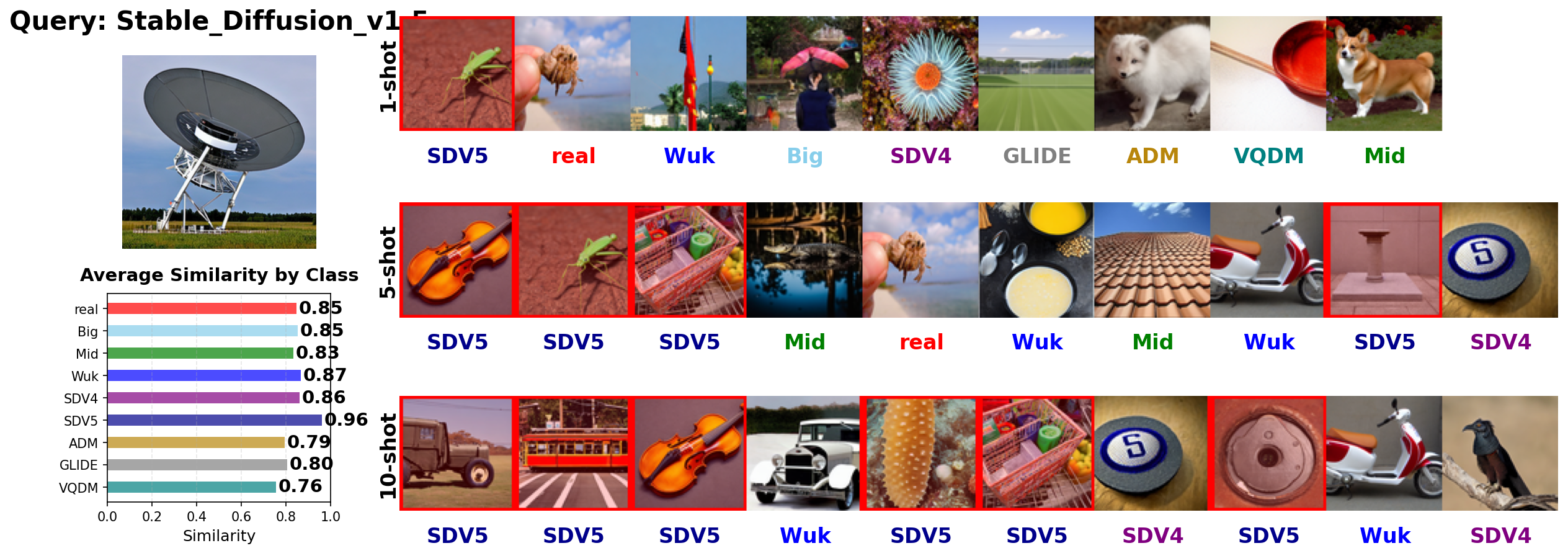}
        \subcaption{Stable Diffusion V1.5 generated query image}
    \end{minipage}
    
    \caption{\textbf{Illustration of retrieval-based image attribution.} The top-ranked retrieval results from the database are listed, with the correct result highlighted by a red bounding box. The queries are a Wukong-generated~\cite{wukong2022} image, a SDV4-generated image, and a SDV5-generated~\cite{stable} image, respectively.}
    \label{fig:visual_retrieval_2}
\end{figure*}

\begin{figure*}[ht]
    \centering
    \begin{minipage}[ht]{\textwidth}
        \centering
        \includegraphics[width=1\textwidth]{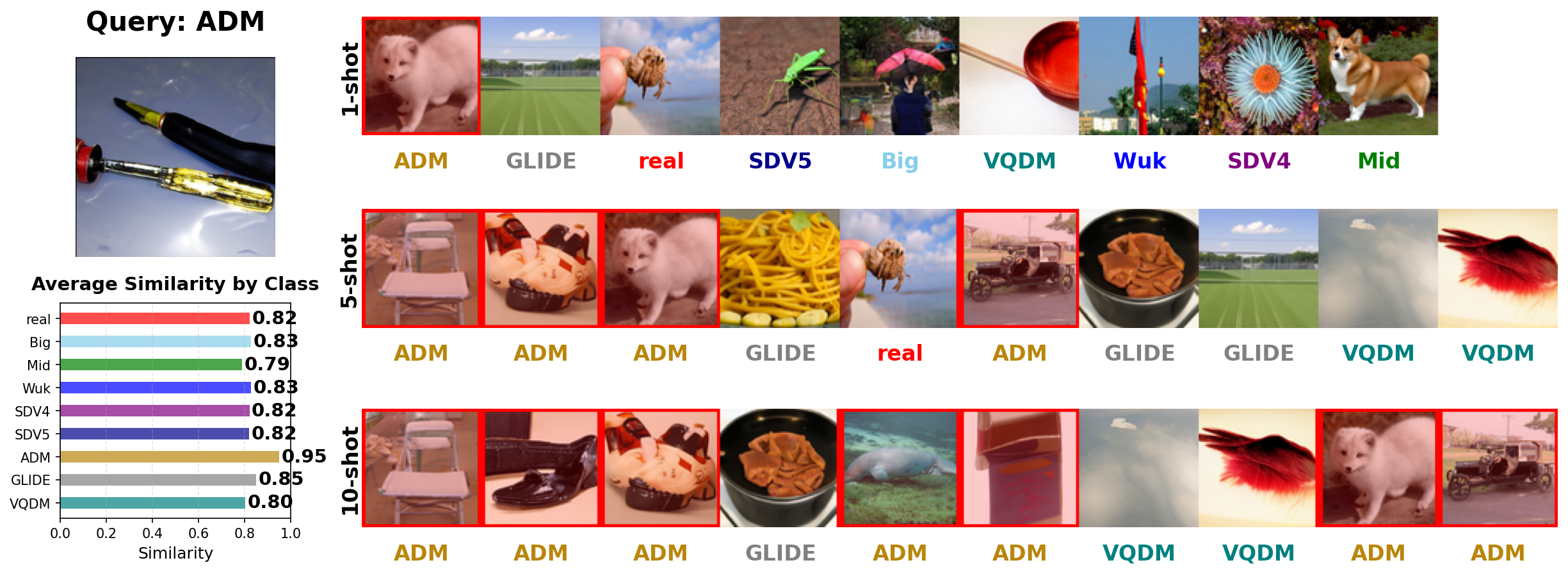}
        \subcaption{ADM generated query image}
    \end{minipage}

    \begin{minipage}[ht]{\textwidth}
        \centering
        \includegraphics[width=1\textwidth]{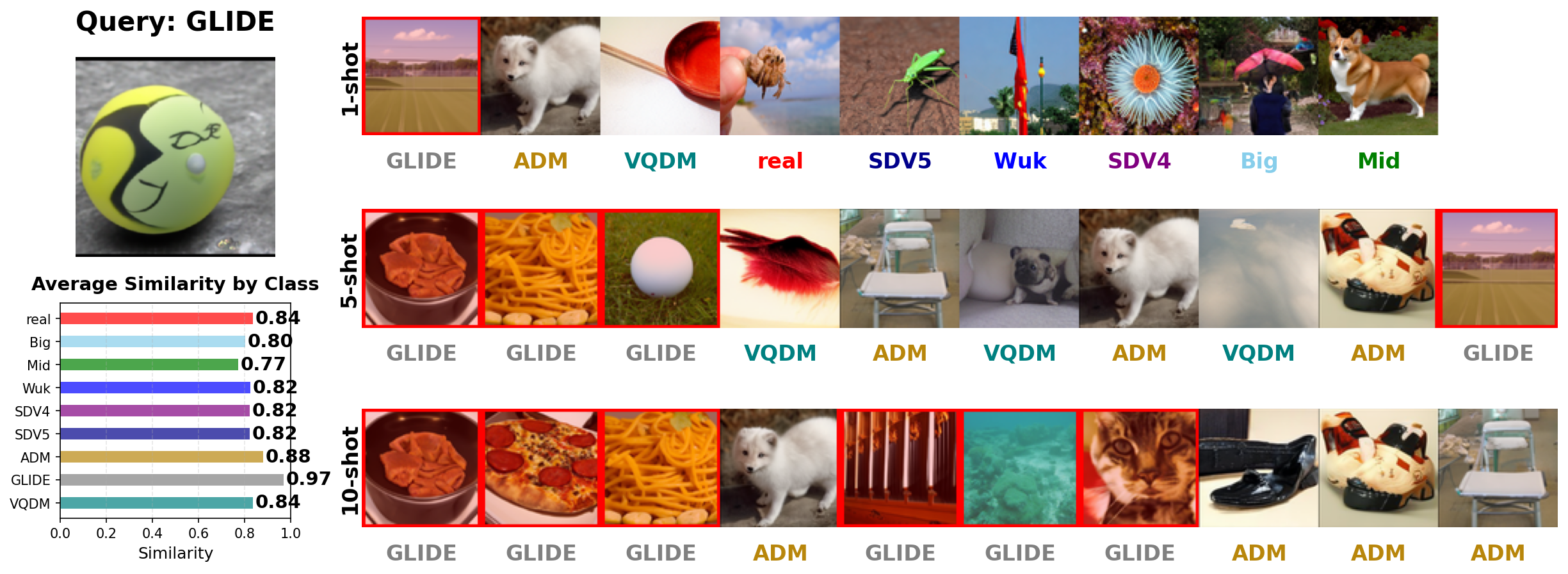}
        \subcaption{GLIDE generated query image}
    \end{minipage}

    \begin{minipage}[ht]{\textwidth}
        \centering
        \includegraphics[width=1\textwidth]{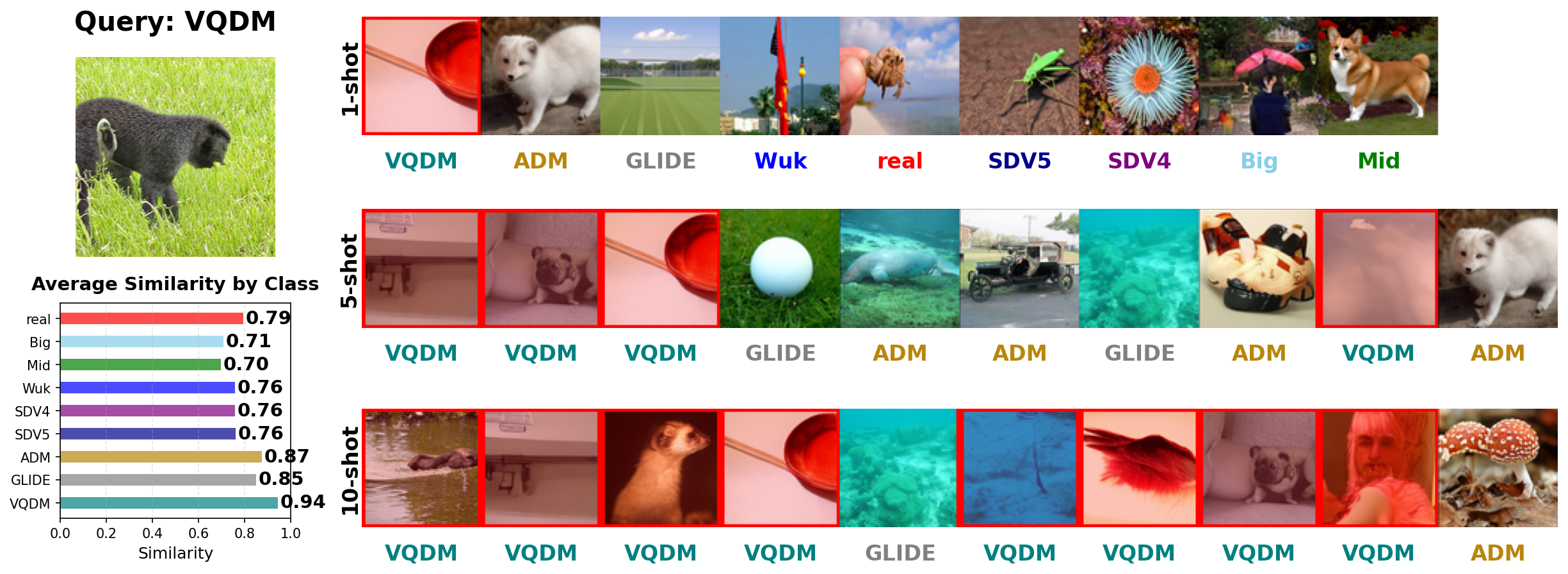}
        \subcaption{VQDM generated query image}
    \end{minipage}
    
    \caption{\textbf{Illustration of retrieval-based image attribution.} The top-ranked retrieval results from the database are listed, with the correct result highlighted by a red bounding box. The queries are a ADM-generated~\cite{adm} image, a GLIDE-generated image~\cite{nichol2021glide}, and a VQDM-generated~\cite{gu2022vector} image, respectively.}
    \label{fig:visual_retrieval_3}
\end{figure*}

\subsection{Visualization}

\noindent{\textbf{Attribution Visualization:}}
We formulate AI-generated image attribution as an instance-retrieval problem, and representative visualization results are shown in Figure~\ref{fig:visual_retrieval_1},~\ref{fig:visual_retrieval_2}, and~\ref{fig:visual_retrieval_3}. 
Specifically, we use a real image and various fake images in GenImage~\cite{zhu2023genimage} dataset as queries, and rank the most similar images from the different-shot database, with the correctly retrieved result highlighted by a red box.
The similarity scores between each query and all images from each generator in the database are visualized using bar charts.
Even though Stable Diffusion V1.4 and Stable Diffusion V1.5~\cite{stable} share a similar core architecture, our method still achieves strong retrieval results.\\

\begin{figure}[ht]
    \centering
    \begin{minipage}[t]{\columnwidth}
        \centering
        \includegraphics[width=\columnwidth]{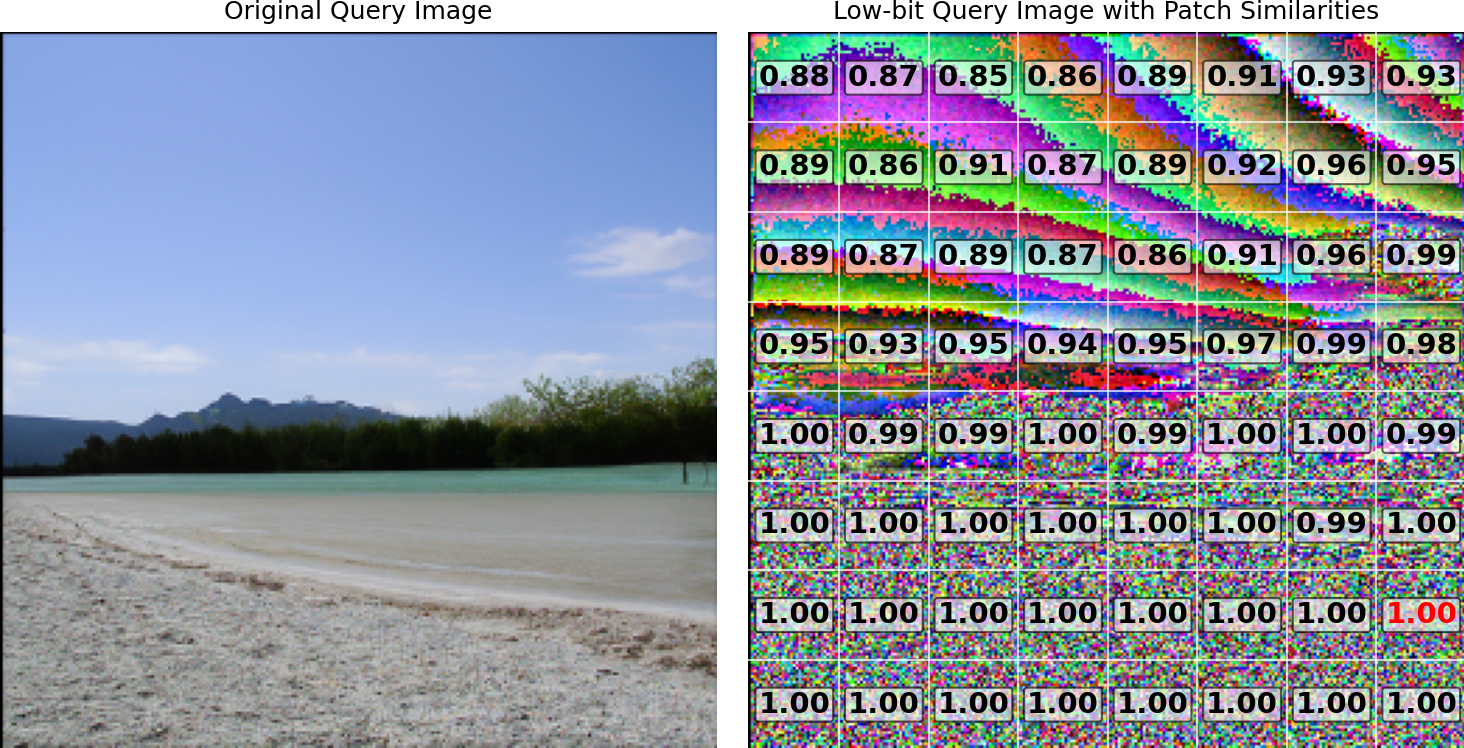}
        \subcaption{}
    \end{minipage}
    
    \vspace{0cm} 

    \begin{minipage}[t]{\columnwidth}
        \centering
        \includegraphics[width=\columnwidth]{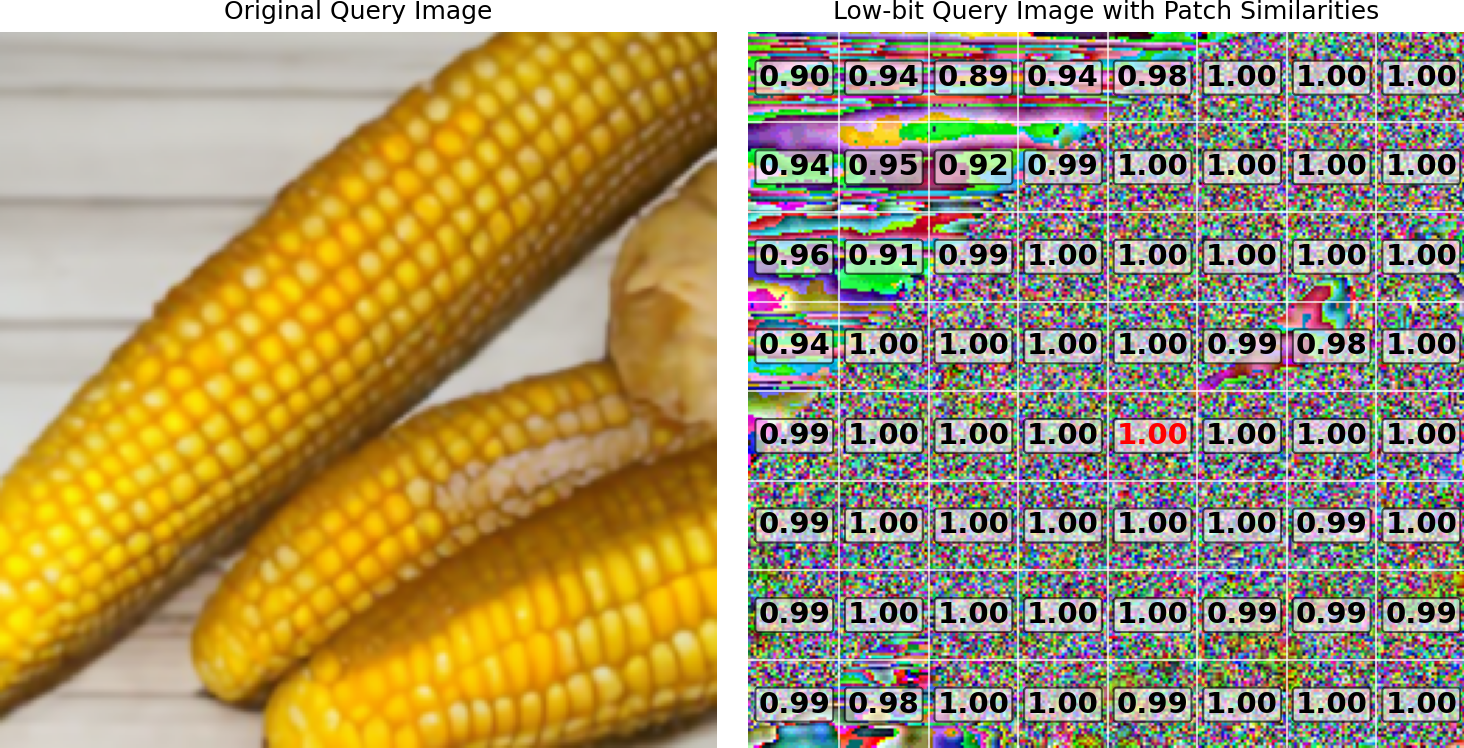}
        \subcaption{}
    \end{minipage}

    \vspace{0cm} 

    \begin{minipage}[t]{\columnwidth}
        \centering
        \includegraphics[width=\columnwidth]{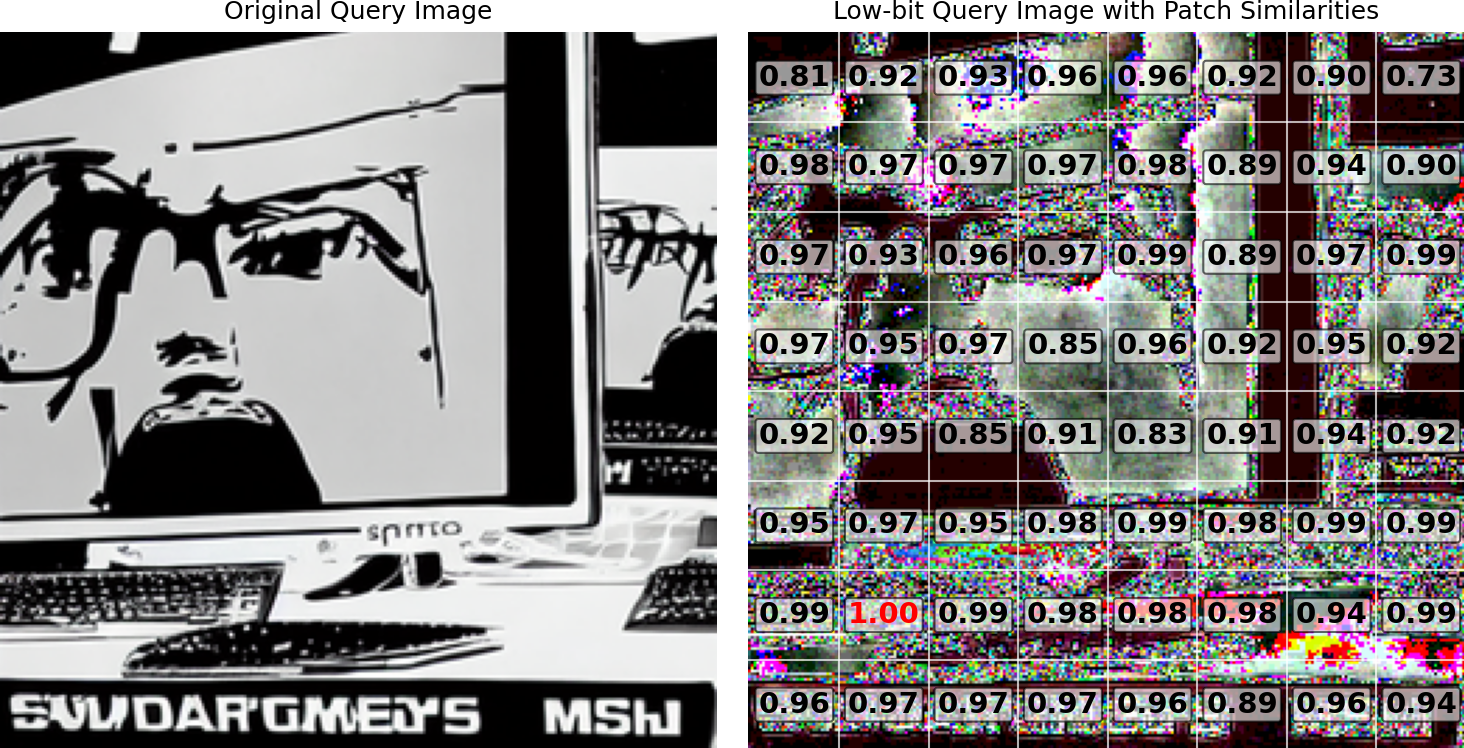}
        \subcaption{}
    \end{minipage}

    \caption{\textbf{Illustration of low-bit generative fingerprints.} Each image is divided into multiple patches, and the patch with the highest similarity score is selected for subsequent processing.}
    \label{fig:visual_fingerprints}
\end{figure}

\noindent{\textbf{Fingerprints Visualization:}}
We visualize the low-bit generative fingerprints of query, which generated by ADM~\cite{adm}, GLIDE~\cite{nichol2021glide} and Wukong~\cite{wukong2022} respectively, as illustrated in Figure~\ref{fig:visual_fingerprints}. In practical implementation, each low-bit generative fingerprint is partitioned into 32×32 patches. For clarity of visualization, we present an 8×8 patch configuration. 
For each query patch, its similarity is computed against every patch of all images in the database, and the maximum similarity is assigned as the score for that patch. 
The patch achieving the highest score is subsequently forwarded to the encoder for feature extraction. This selection process ensures that the features carrying the most significant generator artifacts are used for attribution.

\section{Conclusion}
We propose a versatile and efficient AI-generated image attribution framework called LIDA, which treats attribution as instance retrieval and leverages bit-planes for generative fingerprints extraction. LIDA only trains an attribution encoder using an adapted ResNet-50, and the training involves unsupervised pre-training and few-shot attribution adaptation.
Our forensic technology succeeds in Deepfake attribution and detection under both zero-shot and few-shot settings on two popular AI-generated image datasets. Further experiments demonstrate the effectiveness of each component and the robustness to perturbations and degradations. By relying solely on efficient binary operations and a lightweight encoder, our approach achieves low computational complexity for both training and inference.

{
    \small
    \bibliographystyle{ieeenat}
    \bibliography{main}
}


\end{document}